%% file: main_IMWUT.tex
\documentclass[acmlarge]{acmart}

\usepackage{soul}  
\usepackage{longtable}
\usepackage{tabularx}
\usepackage{tcolorbox} 
\usepackage{subcaption}
\usepackage{enumitem}
\usepackage{color}
\usepackage{colortbl}

\definecolor{LightGray}{gray}{0.85}   
\definecolor{OddGray}{gray}{0.95}     

\AtBeginDocument{%
  }

\setcopyright{none}

\begin{document}

\title{The Promise of Spiking Neural Networks for Ubiquitous Computing: A
Survey and New Perspectives}

\author{Hemanth Sabbella}
\authornote{Both authors contributed equally to this research.}
\email{shrsabbella.2024@phdcs.smu.edu.sg}
\orcid{0000-0002-0889-4657}
\author{Archit Mukherjee}
\authornotemark[1]
\email{architm@smu.edu.sg}
\orcid{0009-0008-3366-7502}
\affiliation{%
  \institution{Singapore Management University}
  \country{Singapore}
}

\author{Thivya Kandappu}
\affiliation{%
  \institution{Singapore Management University}
  \country{Singapore}}
\email{thivyak@smu.edu.sg}
\orcid{0000-0002-4279-2830}

\author{Sounak Dey}
\affiliation{%
  \institution{TCS Research}
  \city{Kolkata}
  \country{India}
}
\email{sounak.d@tcs.com}
\orcid{0000-0002-6199-0437}

\author{Arpan Pal}
\affiliation{%
 \institution{TCS Research}
 \city{Kolkata}
 \country{India}}
\email{arpan.pal@tcs.com}
\orcid{0000-0001-9101-8051}

\author{Archan Misra}
\affiliation{%
  \institution{Singapore Management University}
  \country{Singapore}}
\email{archanm@smu.edu.sg}
\orcid{0000-0003-1212-1769}

\author{Dong Ma}
\authornote{Corresponding author.}
\affiliation{%
  \institution{Singapore Management University}
  \country{Singapore}}
\email{dongma@smu.edu.sg}
\orcid{0000-0003-3824-234X}

\renewcommand{\shortauthors}{Sabbella et al.}

\begin{abstract}
Spiking neural networks (SNNs) have emerged as a class of bio‑inspired networks that leverage sparse, event‑driven signaling to achieve low‑power computation while inherently modeling temporal dynamics. Such characteristics align closely with the demands of ubiquitous computing systems, which often operate on resource‑constrained devices while continuously monitoring and processing time‑series sensor data. 
Despite their unique and promising features, SNNs have received limited attention and remain underexplored (or at least, under-adopted) within the ubiquitous computing community. To address this gap, this paper first introduces the core components of SNNs, both in terms of models and training mechanisms. It then presents a systematic survey of 76 SNN-based studies focused on time-series data analysis, categorizing them into six key application domains. For each domain, we summarize relevant works and subsequent advancements, distill core insights, and highlight key takeaways for researchers and practitioners. To facilitate hands-on experimentation, we also provide a comprehensive review of current software frameworks and neuromorphic hardware platforms, detailing their capabilities and specifications, and then offering tailored recommendations for selecting development tools based on specific application needs.
Finally, we identify prevailing challenges within each application domain and propose future research directions that need be explored in ubiquitous community. Our survey highlights the transformative potential of SNNs in enabling energy-efficient ubiquitous sensing across diverse application domains, while also serving as an essential introduction for researchers looking to enter this emerging field.


\end{abstract}

\ccsdesc[500]{General and reference~Surveys and overviews}

\keywords{Spiking Neural Networks, Ubiquitous Computing, Time-series Signals}


\maketitle
\input{sections/01_Intro_dong}

\input{sections/02_methodology}

\input{sections/03_snn_preliminary}

\input{sections/04_time_series_analysis}

\input{sections/05_hardware_software}
\input{sections/06_future_directions}

\input{sections/07_conclusion}

\bibliographystyle{ACM-Reference-Format}
\bibliography{references}
\end{document}

%% file: sections/01_Intro_dong.tex





\section{Introduction}

Temporal sensor data are fundamental to many ubiquitous computing and robotics applications,  ranging from physiological health monitoring using mobile and wearable devices to robots navigating dynamic environments. Such applications process continuous streams of sensor information to make real-time decisions~\cite{leite2022resource, li2023radar, bock2024temporal, bian2024ubihr}, with the challenge lying in balancing inference accuracy with resource efficiency. For example, a wearable device used for early detection of cardiovascular conditions must process high-frequency data streams while preserving battery life~\cite{seshadri2019wearable}. Similarly, autonomous robots operating in search-and-rescue missions need to analyze dynamic and multimodal inputs, such as LiDAR, cameras, and inertial sensors, while ensuring rapid response time and energy conservation~\cite{sangeeta2023autonomous}. These scenarios illustrate the critical demand for data processing techniques, over streaming sensor data, that are both accurate and computationally efficient, and thus amenable to continuous execution on resource-constrained pervasive devices.

In recent years, machine learning models, particularly deep learning architectures like Recurrent Neural Networks (RNNs), Long Short-Term Memory networks (LSTMs), Convolutional Neural Networks (CNNs), and Transformers, have shown great promise in extracting meaningful patterns from time-series data, effectively mitigating interference from contextual noise~\cite{mohammadi2024deep}. However, these models tend to be computationally intensive and may not be well-suited for real-time inference on resource-constrained devices. In response, researchers have explored various strategies to adapt deep learning models to the resource-limited nature of ubiquitous systems, including model compression techniques (such as pruning~\cite{10.5555/2969239.2969366, ThiNet_ICCV17, molchanov2016pruning, li2016pruning, liu2017learning}, quantization~\cite{han2015deep,polino2018model, wu2016quantized}, and knowledge distillation~\cite{hinton2015distilling, ba2014deep, polino2018model}), lightweight architectures~\cite{howard2017mobilenets,zhang2018shufflenet,tan2019efficientnet}, and low-power optimizations~\cite{9502503, 7460664, Seon_2023_WACV}.

Despite these efforts, conventional Artificial Neural Networks (\textbf{ANNs}) still face significant challenges when it comes to efficiently monitoring context in ubiquitous computing environments. These challenges primarily arise from two key factors: (1) Discrete information processing—ANNs process information in discrete chunks, often aggregating sensor data over specific time periods (referred to as the "detection window") and processing each such chunk independently without considering cross-chunk temporal dependencies. To avoid missed detections, successive windows typically overlap, causing the same data to be processed multiple times, resulting in increased computational and energy overhead.  (2) Continuous-valued representation—ANNs typically process continuous-valued data, usually in the form of floating-point numbers.  This representation, coupled with a large number of model parameters and activations, leads to significant energy consumption and memory overheads.

\begin{figure}[h!]
    \centering
    \includegraphics[width=0.5\linewidth]{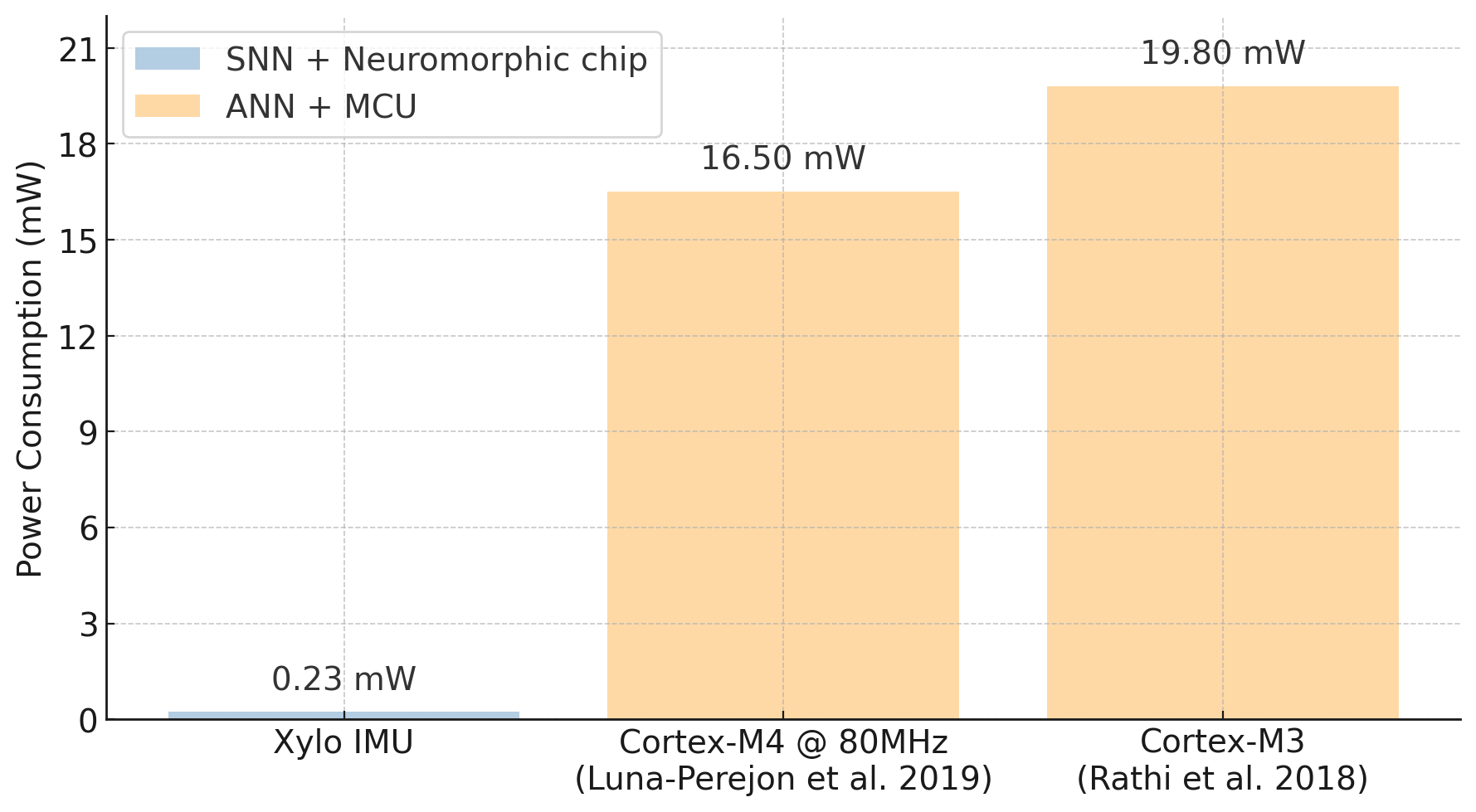}
    \caption{Power comparison of fall detection systems using SNN (on neuromorphic chip) and ANN (on MCU).}
    \label{fig:fall_xylo_mcus}
\end{figure}


In contrast, Spiking Neural Networks (\textbf{SNNs}) offer a fundamentally different, biologically inspired model of neural computation that can help overcome these challenges~\cite{ghosh2009spiking}. Specifically, SNNs simulate the behavior of neurons in human brain by using \textbf{discrete spikes} for communication and computation. Unlike continuous-valued representations, spikes are lightweight, requiring only a single bit of information (1 or 0). Additionally, SNN neurons accumulate information over time and generate spikes only when necessary, reducing both computational load and energy consumption. For instance, as illustrated in Figure \ref{fig:fall_xylo_mcus}, we experimentally found the power consumption for fall detection using an SNN (implemented on neuromorphic hardware) to be significantly lower ($71\times$ reduction) than that of an ANN (implemented on conventional microcontrollers (MCUs)). This event-driven processing paradigm also aligns well with the temporal and dynamic nature of sensor data, where only significant events or changes, occurring sporadically, need to be processed. Moreover, SNNs can naturally encode temporal information, making them particularly well-suited for tasks involving time-dependent data and real-time data streams~\cite{matenczuk2021financial}.


In recent years, SNNs have garnered significant attention for their superior energy efficiency and real-time processing capabilities, positioning them as a promising new avenue for resource-constrained and time-sensitive applications. For example, recent past works have focused on (a) improving SNN \emph{model training} and \textit{architecture design}~\cite{liyanagedera2023low, datta2022towards}
(b) designing and implementing \emph{energy-efficient neuromorphic hardware} aimed at executing SNN models of varying sizes with minimal power consumption~\cite{mehrabi2024real}, 
(c) applying SNN models in various wearable and mobile applications like gesture recognition~\cite{scrugli2024real} and physiological monitoring~\cite{8896021}, for sustained real-time operations on resource-constrained low-end devices.

\definecolor{gold}{RGB}{255,170,10}

\begin{table}[h!]
\centering
\small
\begin{tabularx}{\textwidth}{|p{1.7cm}|p{0.6cm}|p{1.4cm}|p{1.3cm}|p{1.3cm}|p{1.3cm}|p{1.2cm}|p{1.3cm}|p{2.5cm}|}
\hline
\textbf{Survey Paper} & \textbf{Year} & \textbf{Modalities} & \textbf{Neuron Model} & \textbf{Encoding} & \textbf{Training} & \textbf{Software} & \textbf{Hardware} & \textbf{Application Focus} \\
\hline
Catherine et al. \cite{Catherine2017} & 2017 &\begin{tabular}[c]{@{}l@{}}Video \\ Image \\ Time-series \end{tabular}   & \textcolor{gold}{\(\star \star \star \star \star\)} & \textcolor{gold}{\(\star\)} & \textcolor{gold}{\(\star \star \star\)} & \textcolor{gold}{\(\star \star\ \star\)} & \textcolor{gold}{\(\star \star \star\)} & No Focus \\
\hline
Pfeiffer et al. \cite{pfeiffer2018deep} & 2018 & Image & \textcolor{gold}{\(\star\)} & \textcolor{gold}{\(\star\)} & \textcolor{gold}{\(\star \star \star \star\)} & \textcolor{gold}{\(\star\)} & \textcolor{gold}{\(\star \star\)} & Image Classification \\
\hline
Bouveir et al. \cite{bouvier2019spiking} & 2019 & Image & \textcolor{gold}{\(\star \star\)} & \textcolor{gold}{\(\star\)} & \textcolor{gold}{\(\star \star\ \star\)} & \textcolor{gold}{\(\star\)} & \textcolor{gold}{\(\star \star \star\)} & Image Classification \\
\hline
Tavanaei et al. \cite{tavanaei2019deep} & 2019 & Image & \textcolor{gold}{\(\star \star \star\)} & \textcolor{gold}{\(\star\)} & \textcolor{gold}{\(\star \star \star\)} & \textcolor{gold}{\(\star\)} & \textcolor{gold}{\(\star\)} & Image Recognition, Object Detection \\
\hline
Nunes et al. \cite{Nunes2022} & 2022 & Image & \textcolor{gold}{\(\star \star\)} & \textcolor{gold}{\(\star \star\)} & \textcolor{gold}{\(\star \star \star\)} & \textcolor{gold}{\(\star\)} & \textcolor{gold}{\(\star\)} & Image Classification \\
\hline
Yamazaki et al. \cite{yamazaki2022spiking} & 2022 & \begin{tabular}[c]{@{}l@{}}Image 
\end{tabular}  & \textcolor{gold}{\(\star \star \star\)} & \textcolor{gold}{\(\star \star\)} & \textcolor{gold}{\(\star \star \star\)} & \textcolor{gold}{\(\star \star \star\)} & \textcolor{gold}{\(\star\)} & Computer Vision, Robotics \\
\hline
Rathi et al. \cite{rathi2023exploring} & 2023 & \begin{tabular}[c]{@{}l@{}} Image \\ Time-series \end{tabular} & \textcolor{gold}{\(\star \star \star\)} & \textcolor{gold}{\(\star \star \star \star\)} & \textcolor{gold}{\(\star \star \star \star\)} & \textcolor{gold}{\(\star \star\)} & \textcolor{gold}{\(\star \star \star\)} & Image, Gesture, Biomedical, Motion \\
\hline
Eshraghian et al. \cite{eshraghian2023training} & 2023 & Image & \textcolor{gold}{\(\star \star \star \star\)} & \textcolor{gold}{\(\star \star \star \star\)} & \textcolor{gold}{\(\star \star \star \star\)} & \textcolor{gold}{\(\star\)} & \textcolor{gold}{\(\star\)} & Image Classification \\
\hline
Manon et al. \cite{Manon2024} & 2024 & Image & \textcolor{gold}{\(\star \star\)} & \textcolor{gold}{\(\star \star\)} & \textcolor{gold}{\(\star \star \star \star\)} & \textcolor{gold}{\(\star\)} & \textcolor{gold}{\(\star\)} & Image Processing \\
\hline
\textbf{Our Paper} & NA & \textbf{Time-series} & \textcolor{gold}{\( \star \star \star\)} & \textcolor{gold}{\(\star \star \star \)} & \textcolor{gold}{\(\star \star  \star\) } & \textcolor{gold}{\(\star \star \star \star \star\)} & \textcolor{gold}{\(\star \star \star \star \star\)} & \textbf{Ubiquitous Sensing Systems}  \\ \hline
\end{tabularx}
\caption{Comparison of existing SNN surveys. For neuron model, encoding, training, software, and hardware, we use star levels ranging from 1 to 5 stars to represent the comprehensiveness.} 
\label{tab:survey_papers}
\end{table}





Building on these advancements, several papers have systematically reviewed the broader SNN landscape (summarized in Table~\ref{tab:survey_papers}, organized based on comprehensiveness), targeting the machine learning community. Specifically, many surveys offer comprehensive foundational overviews of SNN components, such as neuron models~\cite{Catherine2017,eshraghian2023training}, encoding schemes~\cite{rathi2023exploring,eshraghian2023training}, and training methods~\cite{Manon2024,eshraghian2023training,rathi2023exploring,yamazaki2022spiking,Nunes2022,tavanaei2019deep,bouvier2019spiking,pfeiffer2018deep,Catherine2017}, providing a generic theoretical introduction to researchers interested in the development of SNN. Some surveys also categorizes software~\cite{yamazaki2022spiking, Catherine2017} and hardware~\cite{bouvier2019spiking,Catherine2017,rathi2023exploring} platforms for the implementation of SNNs. However, due to the rapid development of SNN research, some platforms being discussed are now outdated or no longer maintained. Meanwhile, more recent hardware and software developments offer better support for real-time, low-power SNN applications—but have yet to be comprehensively reviewed. 
Additionally, the majority of these surveys primarily focus on image or vision-based data, with only a few, such as \cite{rathi2023exploring, Catherine2017}, briefly touching on time-series data. \textit{None of the existing surveys place a strong emphasis on time-series as a modality, despite the fact that SNNs are particularly well-suited for such data due to their inherent ability to model and exploit temporal dynamics. This oversight is especially limiting in the context of ubiquitous computing, where time-series sensor data is a core modality.} 

Therefore, in this survey, we aim to \textbf{introduce SNNs to the mobile sensing and ubiquitous computing community}, with a specific focus on \textbf{time-series data}. This community, where efficient processing of time-series sensor data is crucial, stand to benefit significantly from the inherent temporal dynamics and energy efficiency of SNNs. To address the deficiencies of existing SNN surveys from the standpoint of ubiquitous applications,  We review 76 peer-reviewed articles spanning various domains, such as computational neurology, contextual sensing, and neuromorphic hardware. We categorize them into different applications based on their relevance to real-world use cases and analyze how SNN-based processing aligns with their requirements. This is further supported by a tailored overview of core SNN components, such as neuron models, encoding schemes, and training methods, specifically adapted for time-series understanding.

Furthermore, we offer actionable guidelines and resources for researchers interested in entering this field, including a comprehensive review of up-to-date software libraries and hardware platforms. This includes a categorized review of recent software toolchains based on the type of support they provide (e.g., simulation/development) and their compatibility with specific hardware platforms. On the hardware side, we categorize available platforms based on their support for network size,
low-power operation, and current maintenance support, offering insight into state-of-the-art neuromorphic chipsets. Finally, we extend our taxonomy of applications with a forward-looking perspective to propose future research directions and opportunities for the ubiquitous computing community.

This paper is organized as follows: Section 2 provides the methodology for literature survey, i.e., specific procedure adapted to list and select the publication corpus. Section 3 introduces the fundamentals of SNNs, including neuron models, encoding techniques, and training strategies relevant to time-series sensor data. Section 4 presents a taxonomy of SNN applications for time-series sensor signals, highlighting both current trends and emerging opportunities. Section 5 surveys the state-of-the-art in software and hardware platforms for SNN development, with a focus on tools suitable for resource-constrained environments.  Finally, Section 6 discusses the future potential of SNNs in ubiquitous sensing systems, outlining key challenges and promising research directions.

%% file: sections/02_methodology.tex
\section{Methodology}




\subsection{Paper Retrieval}


The literature review was curated by gathering papers from various online digital platforms, as neuromorphic computing, especially SNNs, is highly interdisciplinary. We collected papers from IEEE, ACM, MDPI, Frontiers, and Nature publishers. We used \textit{Google Scholar} as the primary search engine, formulating structured search queries that included three key components:

\begin{itemize}
    \item \textit{Application Name} – A specific ubiquitous computing task (e.g., Human Activity Recognition, Emotion Classification).
    \item \textit{Spiking Neural Networks} – To ensure relevance to SNN-based research.
    \item \textit{Sensor Modality} – A modality commonly used for the application (e.g., EEG, EMG, ECG).
\end{itemize}

A typical query was structured as follows: \textit{``[Application Name]'' with ``Spiking Neural Networks'' using ``[Sensor Modality]''}.

\begin{tcolorbox}[colback=lightgray!20,arc=0pt,outer arc=0pt,boxrule=0mm,box align=center] 
\textbf{Some Example Queries} \\ 
\begin{tabular}{l} 
\textit{``Gesture Recognition'' with ``Spiking Neural Network'' using ``sEMG''} \\ 
\textit{``Emotion Classification'' with ``Spiking Neural Network'' using ``EEG''} \\
\textit{``Heart Activity Recognition'' with ``Spiking Neural Network'' using ``ECG''} \\
\end{tabular}
\end{tcolorbox} 




This paper retrieval method resulted in a corpus of 166 papers. The publishers with maximum number of papers are 80 papers in IEEE, 19 papers in Elsevier, 16 papers in Springer, 8 papers in Frontiers, 5 papers in MDPI, 5 papers in ACM Digital Library, 4 papers in IOP Publishing, and 29 papers in other publishers. 






\begin{figure}[t]
    \centering
    \includegraphics[width=0.8\textwidth]{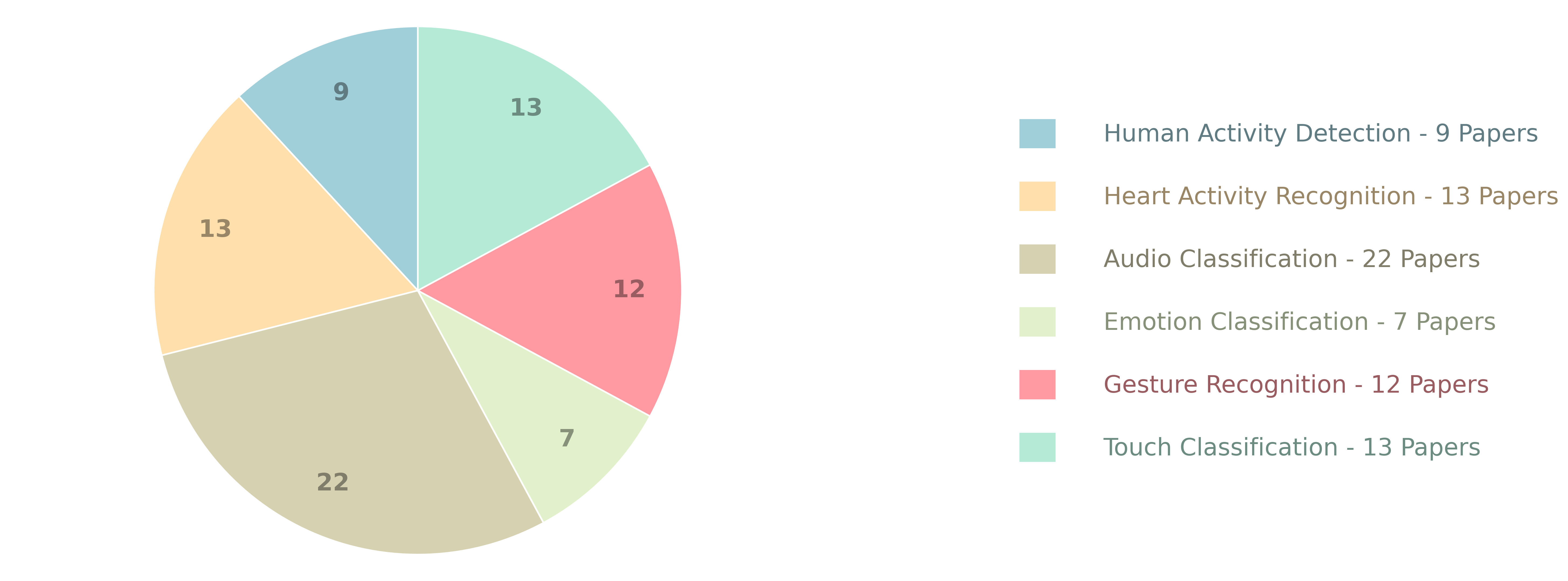}
    \caption{Application-wise distribution of papers in the filtered corpus. } 
    \label{fig:filtered_corpus_pie_chart}
\end{figure}

\subsection{Filtering and Inclusion Criteria}




To ensure the quality and relevance of our review, we applied a two-step filtering process:

\begin{enumerate}
    \item Initial filtering based on relevance
    \begin{itemize}
        \item Papers had to focus on \textit{low-power, real-time implementations} aligned with ubiquitous computing principles.
        \item Papers published in \textit{high-impact venues} or with a \textit{significant citation count} were prioritized.
    \end{itemize}
    
    \item Backward chaining and four-eyed principle 
    \begin{itemize}
        \item The \textit{four-eyed principle} is the process of independent validation by multiple reviewers, where each selected paper is assessed by at least two reviewers to eliminate redundant or less relevant studies.
        \item We performed \textit{backward chaining} by reviewing the reference lists of selected papers to identify additional relevant works, to construct a comprehensive overview of the existing literature.

    \end{itemize}
\end{enumerate}

The filtering produced a total of 76 papers. Further, the filtered corpus can be broken down based on application as follows: 9 papers in Human Activity Detection, 13 papers in Heart Activity Recognition, 22 papers in Audio Classification, 7 papers in Emotion Classification, 12 papers in Gesture Recognition, and 13 papers in Touch Classification  as illustrated in Figure \ref{fig:filtered_corpus_pie_chart}.


%% file: sections/03_snn_preliminary.tex
\section{SNN Preliminary}


In this section, we introduce the fundamentals of Spiking Neural Networks (SNNs), focusing on their foundational principles and how they draw inspiration from biological neurons. We then selectively present commonly used spiking neuron models, particularly those frequently employed in the literature on time-series sensor signals when designing SNN architectures. This is followed by an overview of the most widely used encoding techniques, i.e., methods for representing sensory signals in the spike domain. Finally, we provide a brief explanation of training strategies, the underlying challenges, and how these have evolved over time.

\subsection{Basics}

\begin{figure}[!tbp]
  \centering
    \includegraphics[width=\textwidth]{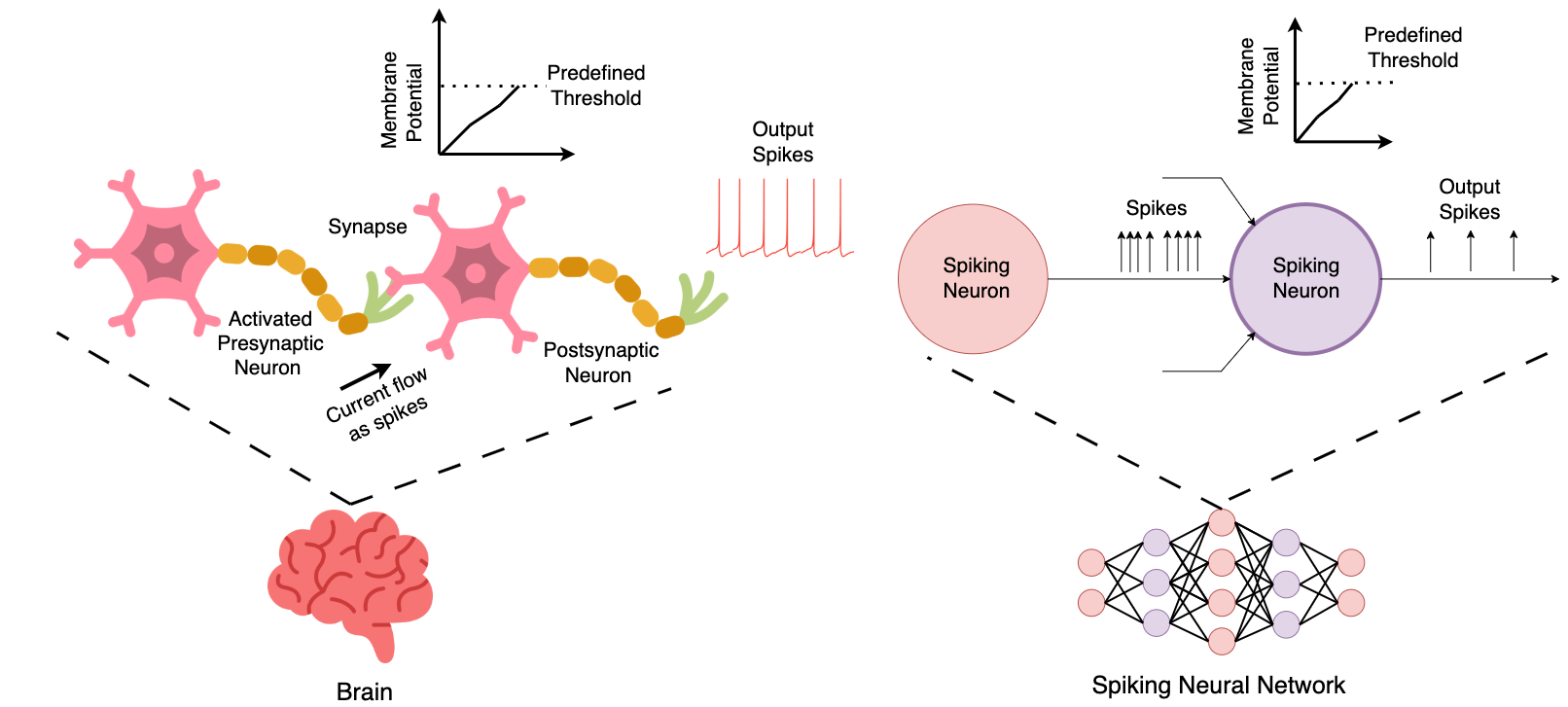}
    \caption{Biological neuron in contrast with Spiking neuron.}
    \label{Biological_Spiking_Neuron}
\end{figure}

SNNs are bio-inspired computational models designed to emulate the spiking behavior of biological neurons~\cite{gerstner2002spiking, gerstner1995time, izhikevich2003simple}. In biological systems, neurons communicate through synapses, where information is transmitted via discrete electrical signals known as spikes. A pre-synaptic neuron transmits the spike information towards the synapse and post-synaptic neuron. 
When a presynaptic neuron activates, it generates a synaptic current that is proportional to the weight of the synapse. This current influences the membrane potential of the connected postsynaptic neuron. Once the membrane potential surpasses a predefined threshold, the postsynaptic neuron generates its own spike and resets its potential to a baseline level. This mechanism is mathematically modeled in SNNs to replicate the functionality of biological neural networks as illustrated in the Figure \ref{Biological_Spiking_Neuron}.

Motivated by this biological phenomenon, SNNs also communicate through discrete spikes, represented as binary signals (0 or 1). These spikes alter the membrane potential of neurons, and once the potential exceeds a threshold voltage, the neuron produces output spikes, resetting the membrane potential. This sparse spike-based communication makes SNNs highly energy-efficient, as neurons only activate when necessary. Moreover, the temporal dynamics inherent in their design make SNNs particularly well-suited for processing event-driven and sequential data, such as time-series signals and sensor outputs~\cite{li, li2, yan2024sparrowsnn, chu2022neuromorphic, bos2024micro, audio3, 9967462, yang2024svad}.

SNNs are built upon three fundamental components. The first is the spiking \textbf{neuron model}, which defines how neurons generate spikes and manage their membrane potentials. These models capture the temporal dynamics that are critical for SNN functionality. The second component is the \textbf{encoding scheme}, which translates input data, such as continuous-valued time-series sensor signals, into spike trains. This allows the network to process information as discrete events. Finally, \textbf{training strategies} are used to optimize the network’s ability to learn and represent spike-based data. These strategies must account for the unique challenges of spike-based communication, such as the non-differentiability of spikes.

In the subsequent sections, we will delve deeper into these components, exploring their types, significance, and the challenges they present in the formulation and practical implementation of SNN models.

\subsection{Spiking Neuron Models}

Spiking neuron models are foundational building blocks for designing SNN architectures. While the architectures of ANNs and SNNs appear to be visually similar, the underlying neuron models are fundamentally different. Spiking neurons communicate asynchronously using discrete spikes, activating only when the membrane potential crosses a certain threshold. In contrast, neurons in ANNs communicate synchronously using a feedforward flow of continuous values. 
This distinction introduces a temporal dependency to computation in spiking neurons, whereas artificial neurons lack time-dependent communication. Moreover, spiking neurons have a state variable, i.e., the membrane potential, which evolves over time based on incoming spikes. ANN neurons, on the other hand, do not have any state variables. This difference enables spiking neurons to retain temporal state information (aka \emph{memory}), making it ideal for sequential data processing while considering temporal correlation over different timescales~\cite{fujii2006temporal, comsa2020temporal, ma2025spiking}.

With these novel features, several spiking neuron models have been mathematically formulated \cite{gerstner2002spiking, izhikevich2003simple, hodgkin1952quantitative, morris1981voltage, ermentrout1996type}. Among them, the Leaky Integrate-and-Fire (LIF) neuron is most widely used in the design of SNNs. This holds true for many applications in ubiquitous computing as well. We will introduce the formulation of LIF neuron and it's alternatives in the following subsections.

\subsubsection{Leaky Integrate-and-Fire (LIF) Neuron~\cite{gerstner2002spiking}}

The LIF neuron is one of the simplest yet most effective representations of biologically plausible neuron models. The LIF neuron's ability to capture essential neural dynamics while remaining computationally efficient is the main reason for its widespread adoption. The LIF neuron has three key functionalities: 

\begin{enumerate}

\item Integration of input: The neuron integrates incoming input spikes over time, causing an increase in input current and raising the membrane potential of the neuron.

\item Leakage: In the absence of input, the membrane potential gradually decays (or "leaks") toward the resting potential, governed by the time constant \( \tau_m \).

\item Fire and reset: If the membrane potential exceeds a predefined threshold \( V_{\text{th}} \), the neuron fires a spike and resets its membrane potential to \( V_{\text{reset}} \).

\end{enumerate}

The mathematical model of the LIF neuron is given by:

\begin{equation}
    \tau_m \frac{dV(t)}{dt} = -V(t) + R I(t)
    \label{lif_neuron}.
\end{equation}

Where:
\begin{itemize}
    \item \( V(t) \): The membrane potential at time \( t \) (in volts, \( V \)).
    \item \( \tau_m \): The membrane time constant (in seconds, \( s \)), which defines the rate at which the membrane potential decays back to the resting potential.
    \item \( R \): The membrane resistance (in ohms, \( \Omega \)), which determines the extent of membrane potential change in response to input current.
    \item \( I(t) \): The input current at time \( t \) (in amperes, \( A \)).
\end{itemize}

\subsubsection{Integrate and Fire (IF) Neuron~\cite{lapicque1907recherches}} 

This neuron model is a subcategory of the LIF neuron, where the LIF neuron is further simplified by removing the leakage mechanism. As a result, the membrane potential in the neuron increases to a large value when a constant input is provided, while it remains static in the absence of input. This model does not account for the biological behavior of neurons, such as the time-dependent decay of the potential. Such simplified models are used when leakage is not significant. The IF neuron model is simply described by the integration formulation:

\begin{equation}
\frac{dV(t)}{dt} = I(t).
\label{if_neuron}
\end{equation}

Where:
\begin{itemize}
    \item \(V(t)\): Membrane potential at time \(t\).
    \item \(I(t)\): Input current at time \(t\).
\end{itemize}

\subsubsection{Recurrent Leaky Integrate and Fire (rLIF) Neuron~\cite{gaospiking}} 
This is a variation of the LIF neuron, where the output spikes of the neurons are fed back into the inputs using feedback connections. Consequently, the membrane potential of the neurons is influenced not only by the input spikes but also by the feedback loop from the neurons themselves. This mechanism allows the network to retain information about past events while processing current input spikes, enhancing the analysis of temporal correlations. While typically implemented using one-to-one recurrence, all-to-all recurrence, where the output spikes of the entire layer are fed back as a weighted sum, has also been proposed. While rLIF neurons can capture complex temporal dynamics, creating an internal memory for sequential processing, they require relatively higher computational demands and longer training periods than LIF neurons. The formulation for the rLIF neuron is given by: 


\begin{equation}
\tau_m \frac{dV_i(t)}{dt} = -V_i(t) + \sum_{j \in \text{presynaptic}} w_{ij} \sum_{t_j^f} \kappa(t - t_j^f) + I_i(t).
\end{equation}

Where:
\begin{itemize}
    \item \( \tau_m \): Membrane time constant (controls the rate of decay of the membrane potential).
    \item \( V_i(t) \): Membrane potential of neuron \( i \) at time \( t \).
    \item \( w_{ij} \): Synaptic weight of the connection from presynaptic neuron \( j \) to neuron \( i \).
    \item \( \kappa(t - t_j^f) \): Synaptic response kernel for spikes from presynaptic neuron \( j \), occurring at firing times \( t_j^f \).
    \item \( I_i(t) \): External input current to neuron \( i \).
\end{itemize}

\subsubsection{Spike Response Model (SRM)~\cite{gerstner1993spikes}} This neuron model is a generalization of spiking behavior, capturing the effects of input spikes and neuronal spiking activity through the membrane potential. It uses predefined kernels to emulate the effect of incoming spikes on the membrane potential and the reset dynamics of the potential after the neuron produces spikes. 
This design is also flexible, allowing its kernels to be modified to resemble LIF or IF neurons. However, this comes at the expense of computational complexity, as it requires tracking spike times and evaluating kernel shapes based on the input spikes. The formulation is as follows:

\begin{equation}
    V(t) = \sum_i \sum_{t_i^f} \eta(t - t_i^f) + \sum_j w_j \sum_{t_j^{\text{in}}} \kappa(t - t_j^{\text{in}}).
\end{equation}

Where:
\begin{itemize}
    \item \( V(t) \): Membrane potential of the neuron at time \( t \).
    \item \( \eta(t - t_i^f) \): Refractory kernel representing the effect of the neuron's own spikes, with \( t_i^f \) being the firing times of the neuron.
    \item \( \kappa(t - t_j^{\text{in}}) \): Synaptic response kernel representing the effect of presynaptic spikes, with \( t_j^{\text{in}} \) being the firing times of the presynaptic neurons.
    \item \( w_j \): Synaptic weight for the connection from presynaptic neuron \( j \).
    \item \( t_i^f \): Times at which the neuron fires spikes.
    \item \( t_j^{\text{in}} \): Times at which presynaptic neurons fire spikes.
\end{itemize}

\textbf{\textit{Remark: }}In this subsection, we focused on the commonly used neuron models based on their relevance in the literature of the SNNs for time-series signals. There are several other classical neuron models, such as the Hodgkin-Huxley~\cite{hodgkin1952quantitative} and Izhikevich models~\cite{izhikevich2003simple}. Hodgkin-Huxley  uses ion channel dynamics between neurons i.e., the membrane allows charged particles to pass in and out, capturing electrical signature of real neurons. On the other hand, Izhikevich abstracts the ion channel behavior and makes it more computationally effective for faster computations. We refer the readers to the relevant literature  \cite{gerstner2002spiking} for a more in-depth study of these complex biologically plausible  models. Additionally, variants of the Leaky Integrate-and-Fire (LIF) model have been widely developed depending on the specific use case—namely, Adaptive LIF, Exponential LIF, Resonate and Fire etc. A more comprehensive review and comparison of different neuron models for SNNs can be found in \cite{Catherine2017, yamazaki2022spiking, eshraghian2023training}.






\subsection{Encoding Techniques}



In typical ANN models, information is computed using floating-point or integer values. In contrast, SNN models rely on spike-based computation, making it essential to convert sensor input data into the spike domain. This necessity has led to the development of various encoding techniques that transform real-valued signals into spike trains. Just as time-series data is represented along the time axis, a spike train is a sequential representation of spikes (either 0 or 1) over time~\cite{auge2021survey}. In this section, we describe the most prevalent encoding schemes used for time-series signals in SNNs and discuss their significance. Additionally, we map the most commonly used encoding schemes to their corresponding ubiquitous sensing modalities, offering a snapshot of where each encoding scheme has been dominantly employed in past work. While mapping, we consider the fundamental principles of the encoding schemes, merging modifications and variants into a single category. We also analyze the rationale behind the preference for specific encoding schemes for particular sensor modality if it is prevalently used. 



\subsubsection{Rate-based Encoding}
\begin{figure}[!h]
  \centering
  \begin{minipage}[b]{0.48\textwidth}
    \centering
    \includegraphics[width=\textwidth]{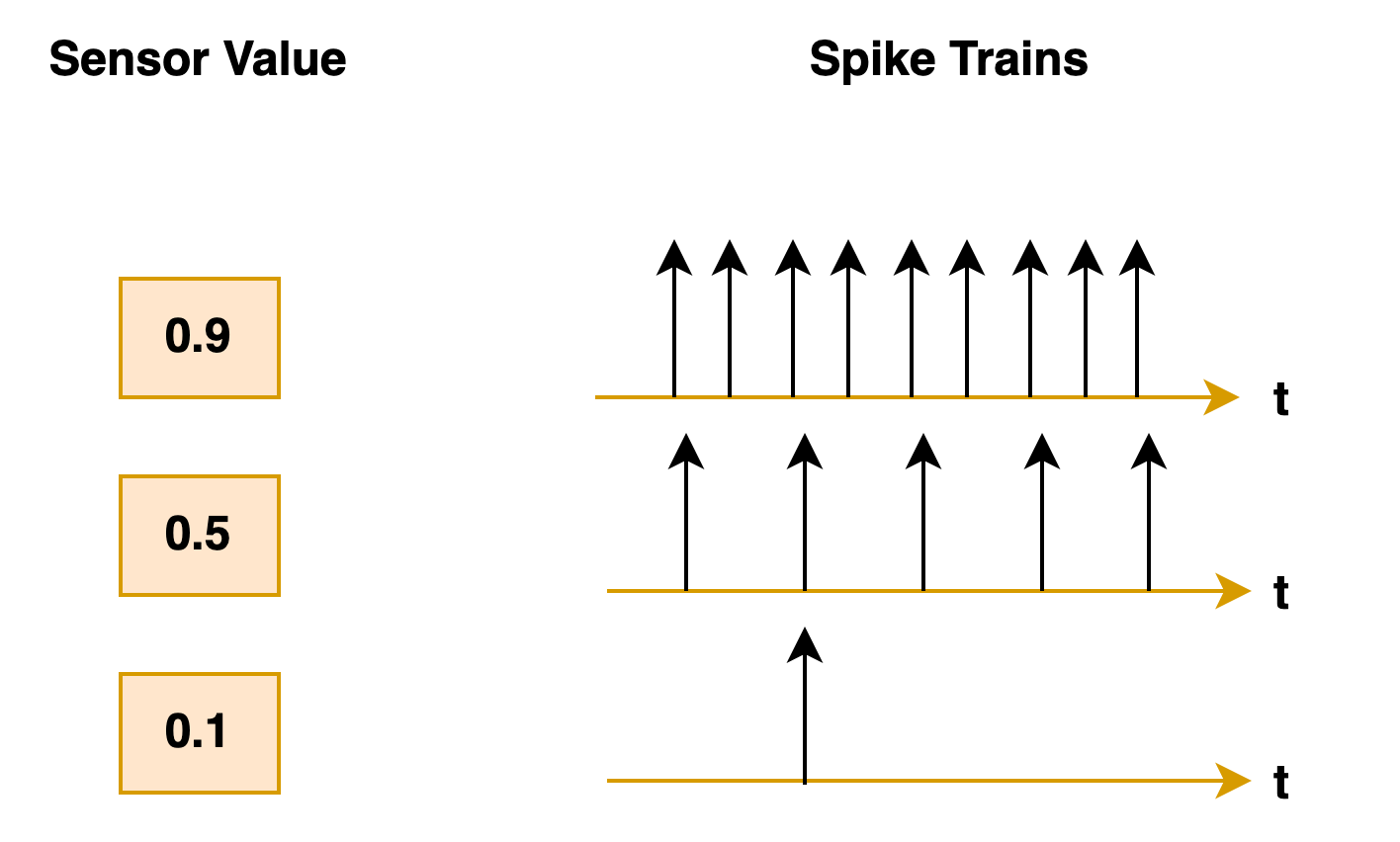}
    \caption{Rate encoding.}
    \label{fig:Rate_encoding}
  \end{minipage}
  \hfill
  \begin{minipage}[b]{0.48\textwidth}
    \centering
    \includegraphics[width=\textwidth]{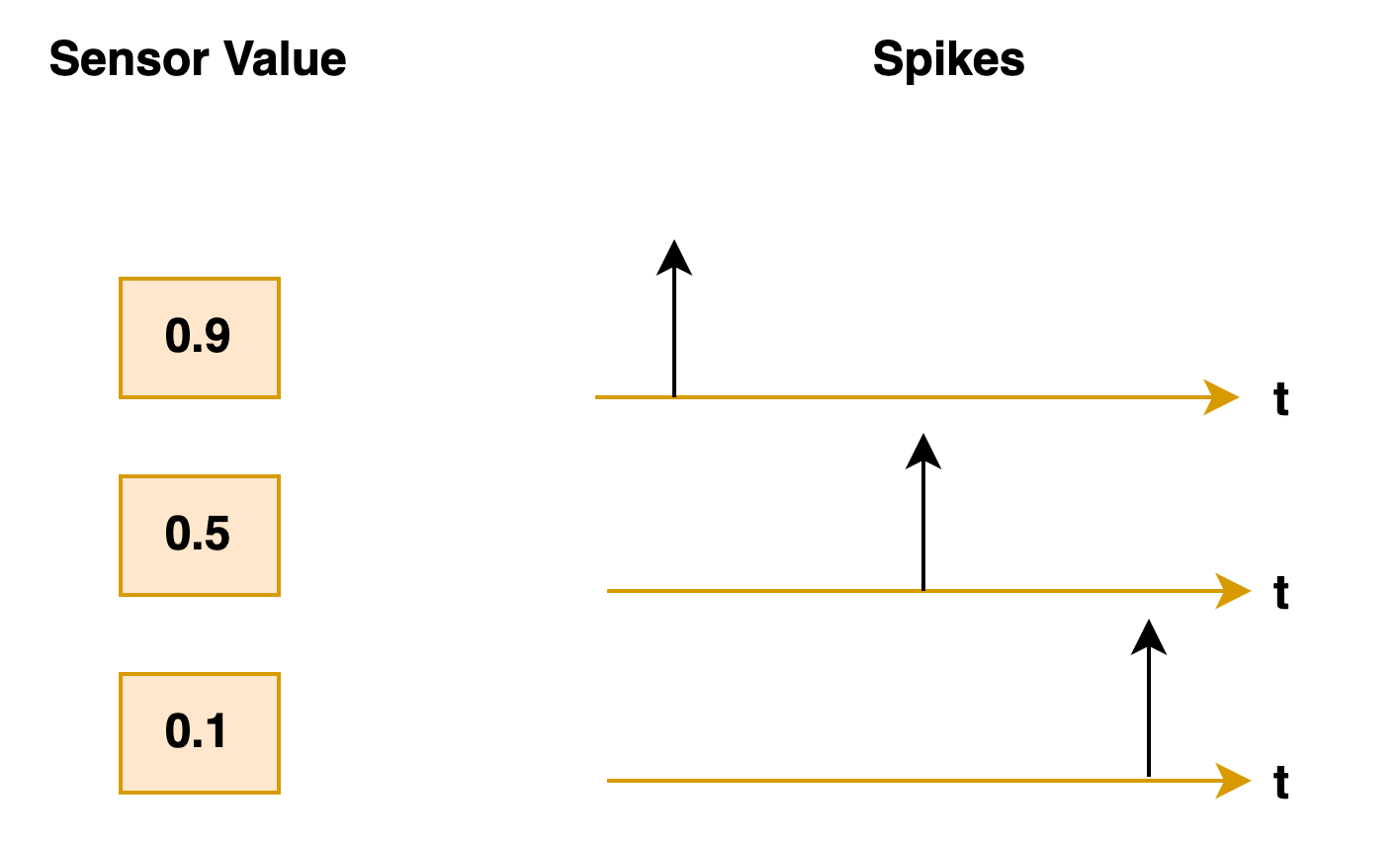}
    \caption{Temporal encoding.}
    \label{fig:latency_encoding}
  \end{minipage}
\end{figure}

Rate-based encoding represents input data as the rate or frequency of spikes, relying on probability distributions to generate spikes. Bian et al. \cite{10.1145/3594738.3611369} defined rate-based encoding using uniform, normal, and beta distributions. In software frameworks like snnTorch, the Poisson distribution is also commonly used to generate rate-based spikes. In rate-based encoding, higher absolute input values correspond to a higher probability of spike generation or a higher rate of spikes in the spike train. For a single sensor data point, rate-based encoding can be represented as either a single spike or a spike train. When using a single spike, the normalized sensor value determines the probability of spike generation. In contrast, a spike train is generated over a defined number of time steps, where the absolute input value defines the probability of spikes across the entire spike train as shown in the Figure \ref{fig:Rate_encoding}. 

\textbf{\textit{Remark: }}Rate-based encoding is predominantly used for neuromorphic image data because the higher number of spikes in rate encoding promotes more learning in SNNs~\cite{diehl2015fast, sengupta2019going, lee2019enabling, shrestha2018slayer, tavanaei2019deep}. This is due to the high spiking activity, which provides a rich representation of spatial features in the spike trains. Rate-based encoding has also been utilized by several works in sensing applications such as human activity recognition (HAR), heart sounds, gesture and other audio analysis \cite{suriani2021smartphone, feng2022building, 8879613, yan2021energy, 10.1016/j.ins.2021.11.065, echoWrite, stereoGest, tieck2020spiking, 10.1145/3625687.3625788, mukhopadhyay2022acoustic}. In two interesting works, Banerjee et al.~\cite{UnivariateMI, MultivariateMI} have proved that embedding Gaussian noise,
into the input time series data, helps increase the overall information content
of the rate encoded spike train, further improving the subsequent SNN performance.





\subsubsection{Temporal Encoding}
Temporal encoding focuses on the timing of spikes, where information is encoded into the specific instances when spikes occur. In time-to-first-spike encoding, higher absolute values of the signal result in earlier spikes, while lower absolute values generate later spikes, which is illustrated in the Figure \ref{fig:latency_encoding}. The timing of the spikes is inversely proportional to the sensor values. This approach produces fewer spikes per data point within a given time window compared to rate-based coding, making it more energy-efficient as fewer spikes consume less dynamic power~\cite{han2020deep, datta2021training}. However, this comes at the cost of representing information with only a single spike, making it more prone to errors and less robust compared to the rate-based representation. There are other temporal encoding schemes, such as gaussian temporal encoding~\cite{TSClassification, TSForecasting, AFClassification}, in which each input is projected into a set of gaussian receptive fields and converted into probabilistic spikes across a population of neurons. This approach essentially divides the input range into several overlapping regions each managed by it's own neuron. When a sample arrives, its value is compared against every zone to determine how likely each neuron is to fire within a short burst of time. This results in a compact spike representation across the population that captures subtle differences in the input, providing a rich spatio-temporal encoding.

\textbf{\textit{Remark: }} There are several works in time series sensor signals explored temporal encoding, specifically in HAR, audio and emotion application \cite{10.1145/3594738.3611369, tan2021brain, yang2018real, dong2018unsupervised, martinelli2020spiking, luo2020eeg}. Recent studies ~\cite{TSClassification, TSForecasting, AFClassification} have also shown how gaussian temporal encoding for various time series data can result in higher performance, compared to rate encoding.


\subsubsection{Delta Encoding}
Delta encoding leverages delta change in input sensor signals to generate spikes~\cite{miskowicz2006send}. Delta encoding is categorized into two types defined as follows:
\begin{itemize}[leftmargin=*]
    \item \textbf{Delta:} In simple delta modulation, a spike is produced when any two subsequent input values change by more than a predefined delta. This can produce positive and negative spikes to reflect the direction of change.
    \item \textbf{n-Threshold Delta:} n-Threshold Delta defines multiple predefined delta values corresponding to the degree of change in the input. These deltas produce $n$ spike trains where each spike train corresponds to an input change that crosses the corresponding delta threshold~\cite{bian2024evaluation}.
\end{itemize}

\begin{figure}[!h]
  \centering
    \includegraphics[width=0.8\textwidth]{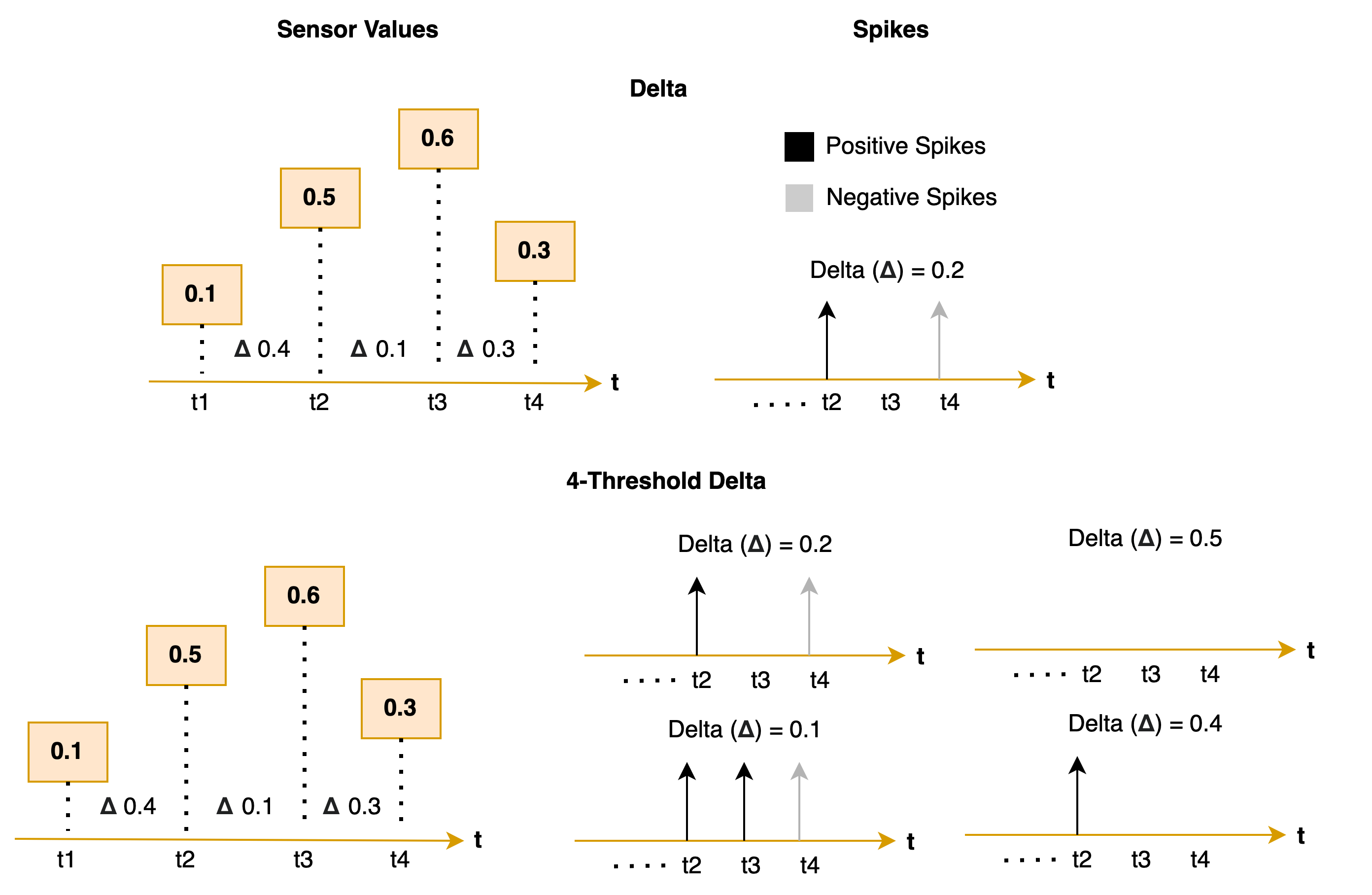}
    \caption{Delta encoding.}
    \label{fig: delta_encoding}
\end{figure}

Figure \ref{fig: delta_encoding} illustrates both types of delta modulation. As shown in the specific example, in the case of basic delta modulation with a predefined delta value of 0.2, whenever the change in the input exceeds this threshold, a spike is generated at those specific time steps—positive spikes are indicated in black, and negative spikes in grey. Similarly, in the case of n-Threshold Delta encoding, four predefined delta values are used (i.e., 0.1, 0.2, 0.4, 0.5). When the input change exceeds any of these thresholds, a spike is generated in the corresponding spike train. These spike trains are typically fed as different channels when processed by SNNs.

\textbf{\textit{Remark: }}Several works have utilized delta modulation, particularly in the applications of HAR \cite{10.1145/3594738.3611369, bian2024evaluation}, heart activity \cite{9948627, ecgIJCNN2019, 8896021}, and gesture recognition \cite{8747378, ceolini2020hand, 9849452, 9073810, scrugli2024real}. Due to its fundamental nature of encoding changes in input signals, delta modulation is especially useful for movement-based classification tasks using IMU data (e.g., detecting anomalies or falls through sharp changes), irregular heartbeats in heart activity monitoring, and motion variations in gesture recognition. Specifically, gesture-based studies have employed delta modulation and demonstrated superior performance~\cite{scrugli2024real, 9849452, 9073810}.


\subsubsection{Level Crossing (LC) Sampling}
LC Sampling is another type of spike encoding that utilizes input signal transitions between quantized levels~\cite{moser2014stability}. The number of levels is predefined, and spikes occur whenever the input crosses from one level to another. While this method also captures changes in the input signal, it differs from delta encoding in that the thresholds are uniformly quantized and fixed. This method effectively captures the signal's transitions.

\begin{figure}[!h]
  \centering
    \includegraphics[width=0.7\textwidth]{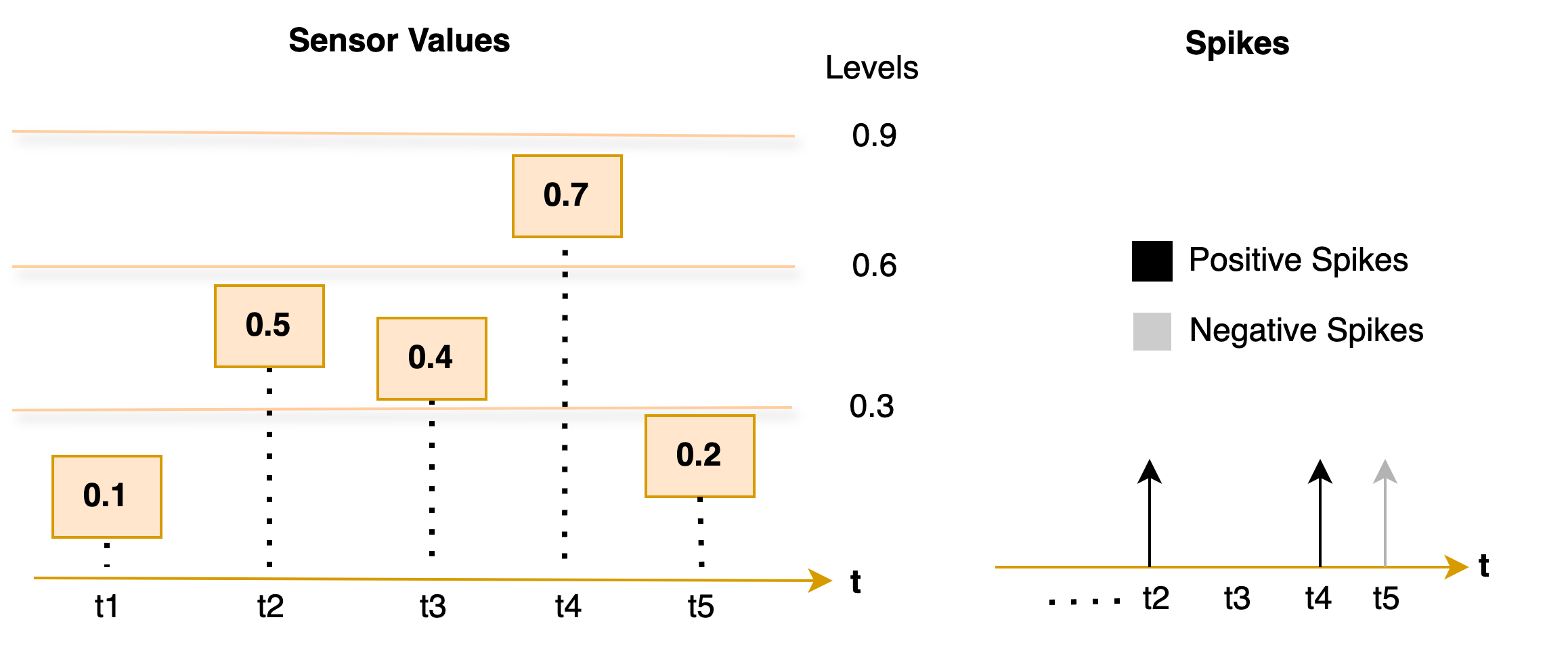}
    \caption{Level Crossing (LC) Sampling.}
    \label{fig: lc_sampling}
\end{figure}

Figure \ref{fig: lc_sampling} illustrates how spikes are encoded using LC Sampling. There are uniformly spaced predefined levels—0.9, 0.6, and 0.3. When the input signal crosses from one level to another (e.g., from time step t1 to t2, the signal goes from 0.1 to 0.5, crossing the 0.3 level), a spike is generated in the spike train. Similarly, from t4 to t5, the input drops and crosses the 0.6 and 0.3 levels, resulting in a negative spike (shown in grey). Similar to n-Threshold Delta modulation, LC Sampling can also utilize multiple thresholds, each with its corresponding spike train. These spike trains can be fed into SNNs as different channels or input streams. 

\textbf{\textit{Remark: }}LC sampling is also used by few works in the heart and touch applications \cite{chu2022neuromorphic, 9644939, taunyazov2020fast}. Specifically, heart activity is coherent with LC sampling as the maximum and minimum values of the heart signal are often consistent, and using a fixed threshold can help to determine subtle changes well. 

\subsubsection{Direct Encoding}

Instead of manually converting sensor values into the spike domain, direct encoding uses the first layer of neurons in an SNN model, or a small SNN network, to encode input sensor signals~\cite{rathi2021diet}. Specifically, Leaky Integrate-and-Fire (LIF) neurons and their variants, such as Integrate-and-Fire (IF) neurons, are commonly used to generate spike representations in the direct encoding. The first layer receives analog sensor values as input and converts them into spike representations across different channels using LIF or IF neurons. The encoding process is also parameterized, meaning the neurons integrate weighted input values and generate spikes at each timestep based on their membrane potential dynamics. Due to its alignment with the fundamental behavior of biological neurons, the LIF-based encoding layer has been widely used in time-series data processing.

\textbf{\textit{Remark: }}Numerous works have adopted this traditional form of spike encoding across domains including HAR \cite{li, li2, fra2022human, shen2021dynamic}, heart activity \cite{yan2024sparrowsnn, rana2024electrocardiography}, audio classification \cite{10.1145/3320288.3320304, weidel2021wavesenseefficienttemporalconvolutions, wu2020deep, leow2023sparsity, wu2021hurai}, gesture recognition \cite{guo2024spgesture}, touch sensing \cite{9665453, follmann2024touch, 9937733}, and emotion recognition \cite{xu2024eescn, yan2022eeg}. This is the most widely used form of spike encoding due to its biological neuron behavior, compatibility with neuromorphic hardware, and its ability to preserve rich temporal dynamics in time-series signals.

\subsection{Training Procedures}

SNNs differ fundamentally from traditional ANNs due to their spiking nature and rich temporal dynamics, therefore necessitating dedicated training strategies. 
Specifically, the binary nature of spikes and the non-differentiability of activation functions can lead to some neurons failing to spike during training, resulting in zero gradients during backpropagation—an issue known as the \textbf{dead-neuron problem}. As shown in the equation below,




\begin{equation}
    S = \delta(V - V_{th}).
\end{equation}

\begin{equation}
   \frac{\partial S}{\partial V} = 
\begin{cases}
    \infty,         & V = V_{th}\\
    0,              & V \neq$ $V_{th}
\end{cases}
\end{equation}

Where,

\begin{itemize}
    \item $S$: The spike output.
    \item $V$: The membrane potential.
    \item $V_{th}$: The threshold voltage.
    \item $\delta$: Dirac-delta function.
\end{itemize}

\begin{figure}[ht]
    \centering
    \begin{subfigure}[t]{0.68\linewidth}
        \centering
        \includegraphics[width=\linewidth]{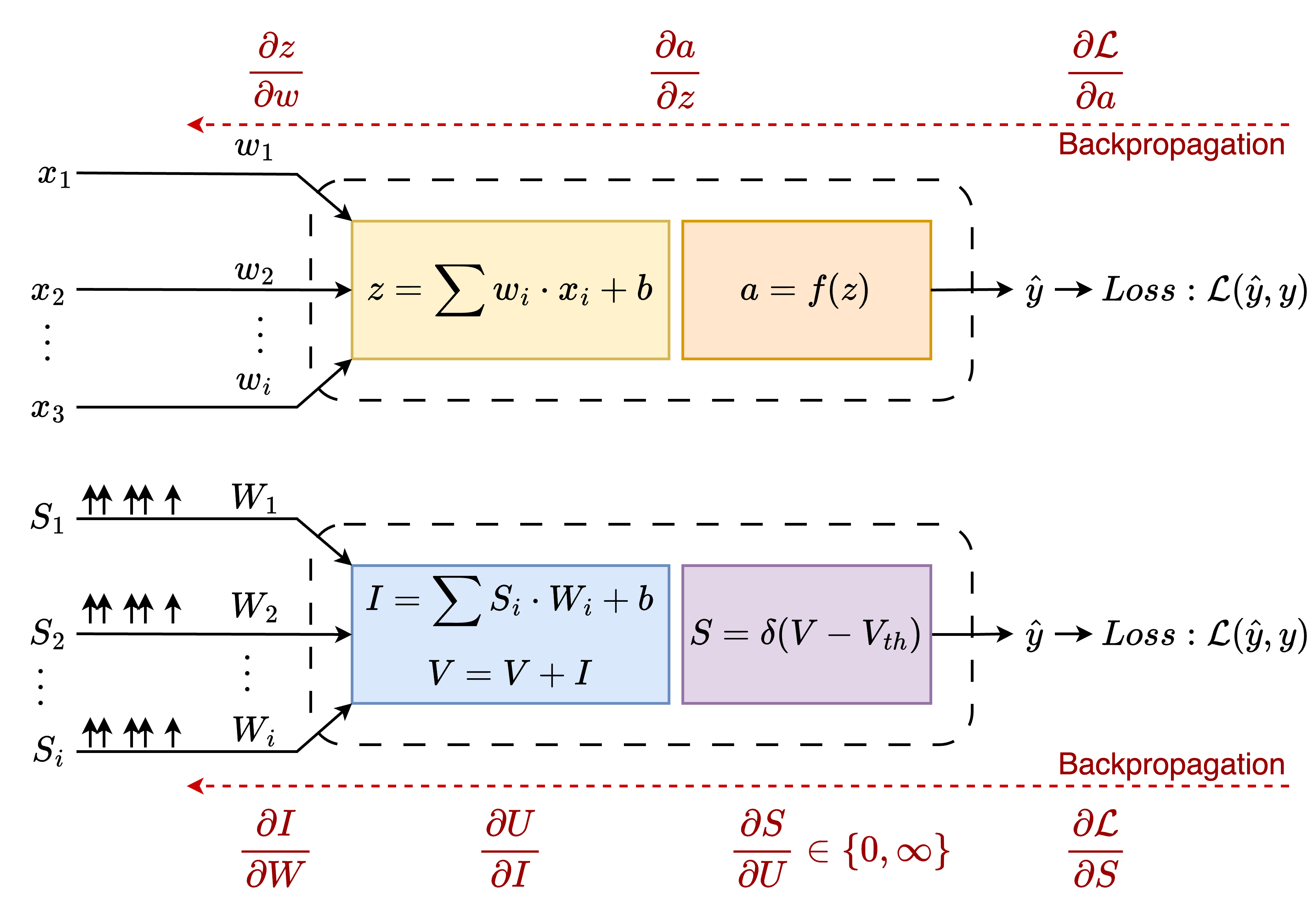}
        \phantomsubcaption
        \label{fig:backprop_compar_a}
    \end{subfigure}
    \hfill
    \begin{subfigure}[t]{0.28\linewidth}
        \centering
        \includegraphics[width=\linewidth]{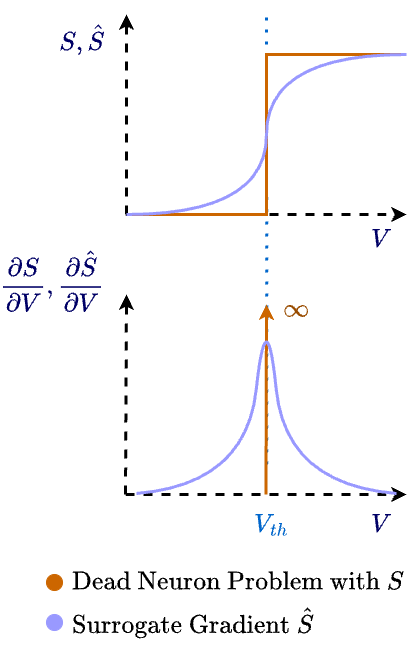}
        \phantomsubcaption
        \label{fig:backprop_compar_b}
    \end{subfigure}
    \caption{(a) Dead neuron problem in the flow of gradients in ANN (top) vs SNN (bottom) backpropagation (left), (b) The dead neuron problem and surrogate gradient descent (right).}
    \label{fig:backprop_compar }
\end{figure}

A spike activity $S$ occurs when the membrane potential $V$ crosses the threshold voltage $V_{th}$. A comparison between the flow of error signal in backpropagation based training in ANN and SNN is illustrated in Figure \ref{fig:backprop_compar_a}. To address this issue, a widely used and intuitive approach is to approximate the non-differentiable spike function with a smooth, differentiable surrogate function during the backward pass~\cite{zenke2018superspike, neftci2019surrogate}, as illustrated in Figure~\ref{fig:backprop_compar_b}. 


In this section, we briefly review the most commonly used training mechanisms for SNNs based on surrogate gradients. For a more detailed discussion, we refer readers to \cite{yamazaki2022spiking, rathi2023exploring, eshraghian2023training, Manon2024}.


\subsubsection{Spike Time Dependent Plasticity (STDP)}
STDP is an unsupervised learning technique where the weights between two neurons are changed based on the spike activity timing \cite{song2000competitive}. The weight updates are triggered only when the pre- and post-synaptic neurons spike within a small time window are strengthened or weakened depending on the temporal order of spiking. If a pre-synaptic spike precedes a post-synaptic spike, the connection is strengthened (i.e., LTP - Long-Term Potentiation). If a post-synaptic spike precedes a pre-synaptic spike, the connection is weakened (i.e., LTD - Long-Term Depression).
Equation \ref{stdp_weight_updation} as given in \cite{song2000competitive} further elaborates upon the weight update rule:
\begin{equation}
    \Delta w = \begin{cases} A_- e^{\Delta t / \tau_-} & \text{if } \Delta t < 0 \\ A_+ e^{-\Delta t / \tau_+} & \text{if } \Delta t > 0  \end{cases}
    \label{stdp_weight_updation}
\end{equation}

Where,

\begin{itemize}
    \item $\Delta t$: The time difference between pre and post synaptic spikes.
    \item $\tau_+$ and $\tau_-$: Time constants for weight update.
    \item $A_+$ and $A_-$: These parameters govern the maximum possible change.
\end{itemize}

The time difference between pre and post synaptic spikes is given by $\Delta t = t_{pre}-t_{post}$  , normally in the range of 10ms, and $\tau_-$ and $\tau_+$ are the time-constants. The parameters $A_+$ (>0) and $A_-$ (<0) need not be fixed and can vary throughout the network from synapse to synapse. 



While this Hebbian based learning rule is commonly used for STDP, there has been different approaches indluding anti-Hebbian STDP (aSTDP) \cite{bell1997synaptic} that updates the weights with opposite dependence on the temporal order of spiking, MirroredSTDP (mSTDP) \cite{burbank2015mirrored} that combines both STPD and aSTPD, probabilistic STDP \cite{masquelier2007unsupervised} that performs LTP proportional to the exponential of current weight, and reward modulated STDP (R-STDP) \cite{izhikevich2007solving} which implements reinforcement learning mechanism for learning.

\subsubsection{SpikeProp}
SpikeProp, proposed by Bohte et al. in 2002 \cite{bohte2002error}, was among the first supervised learning algorithms to adapt backpropagation for SNNs. The method uses spike timing as a continuous differentiable learning signal to align actual spike times with target spike times. This avoids the "dead neuron" problem by leveraging the differentiability of time rather than the non-differentiable spike output.
SpikeProp replaces the derivative of the spike function $\partial S / \partial U$ with the derivative of spike timing $\partial t/ \partial U$, enabling effective weight updates.  

Although SpikeProp showed good performance, it had some limitations. Fundamentally, it assumes that each neuron fires only once, limiting its applicability to multi-spike tasks and dynamic input streams. Moreover, due to its dependency on the occurrence of a spike for weight update, neurons that do not spike during training receive no weight updates, which can hinder learning in deeper networks. Despite these challenges, SpikeProp paved the way for gradient-based SNN training, inspiring more advanced methods.


\subsubsection{Backpropagation Through Time (BPTT)}
BPTT is a gradient-based learning method motivated by the training mechanisms of ANN-based sequential models such as LSTMs, RNNs, and GRUs, which unfolds the temporal dynamics of an SNN into a series of static layers~\cite{werbos1990backpropagation}. This allows errors to propagate backward through time, capturing the temporal dependencies of spike patterns. Unlike methods like SpikeProp, which rely on spike timings for weight updates, BPTT computes gradients over a temporally unrolled computational graph. This ensures that weight updates are not limited to neurons that spike during a specific forward pass.

The gradient of the loss function with respect to a learnable weight is expressed as:

\begin{equation}
\frac{\partial \mathcal{L}}{\partial W} = \sum_{t}\frac{\partial \mathcal{L}[t]}{\partial W} = \sum_{t}\sum_{s\leq t}\frac{\partial \mathcal{L}[t]}{\partial W[s]}\frac{\partial W[s]}{\partial W}
\end{equation}

Where,
\begin{itemize}
    \item $\mathcal{L}[t]$: The instantaneous loss calculated at time step $t$.
    \item $W$: Represents the learnable parameters.
    \item $W[s]$: The instance of the weight applied at previous time steps $s(<t)$.
\end{itemize}

The summation accounts for the temporal dependencies, ensuring that all relevant weight instances contribute to the gradient. This approach has emerged as one of the most popular methods for supervised SNN training due to its flexibility and effectiveness in handling complex temporal patterns.

\subsubsection{Spike Layer Error Reassignment (SLAYER)}
SLAYER is a framework proposed by Shrestha et al. in \cite{shrestha2018slayer} that extends BPTT with a focus on the temporal dependency between the input and the output of a spiking neuron. Similar to how the conventional backprop approach distributes error back through an ANN's layers, SLAYER distributes the credit of error back through the SNN layers. However, because the current state of a spiking neuron depends on its past states (and thus on the prior states of its input neurons), SLAYER also distributes the credit of error back in time, unlike backprop. As a result, it has the ability to simultaneously learn axonal delays and synaptic weights.

\subsubsection{Spatio-Temporal Backpropagation (STBP)}
STBP, proposed by Wu et al. in \cite{wu2018spatio}, is another significant training method for SNNs, addressing the challenges by combining spatial and temporal dependencies in learning. The gradients are computed and flowed backward iteratively for both spatial and temporal domains, ensuring comprehensive weight updates as shown in the following equation:
\begin{equation}
\frac{\partial \mathcal{L}}{\partial W[n]} = \sum_{t}\frac{\partial \mathcal{L}[t]}{\partial W[n]} = \sum_{t}\sum_{s\leq t}\frac{\partial \mathcal{L}[t]}{\partial u[s,n]}\frac{\partial u[s,n]}{\partial I[s,n]}\frac{\partial I[s,n]}{\partial W[n]}
\end{equation}

Where,
\begin{itemize}
    \item $s$: The previous time steps.
    \item $n$: Represents the layer index.
    \item $\mathcal{L}[t]$:The loss at time step $t$.
    \item $W[n]$: Weight at layer $n$.
    \item $u[s,n]$: Membrane potential at layer $n$ at time step $s$.
    \item $I[s,n]$: The input current to the layer $n$ at time step $s$.
\end{itemize}

Unlike methods that primarily focus on either spatial domain (SD) or temporal domain (TD) features, STBP integrates both by propagating errors through the spatial layers and across temporal dynamics, thereby fully leveraging the spatio-temporal information encoded in spikes.

\subsubsection{Supervised Spike Time Dependent Plasticity (SSTDP)}
SSTDP, proposed by Liu et al. in \cite{liu2021sstdp}, is another popular training mechanism that combines the biological plausibility of STDP with supervised learning objectives to achieve efficient training of SNNs. SSTDP introduces a hybrid learning approach that integrates local weight updates (STDP) with global error signals (backpropagation). This combination ensures both efficient feature extraction and effective global optimization. Like STDP, temporal encoding is employed for SSTDP to better mimic the working of the brain. Likewise, the loss function in SSTDP is based on the temporal discrepancy between the actual firing times and the expected firing times of neurons. The gradients are backpropagated according to the equation below:

\begin{equation}
\frac{\partial \mathcal{L}[n]}{\partial W[n]} = \frac{\partial \mathcal{L}[n]}{\partial t[n]}\frac{\partial t[n]}{\partial W[n]}.
\end{equation}

Where,
\begin{itemize}
    \item $\mathcal{L}[n]$: The squared error of the difference between actual firing time and expected firing time.
    \item $t[n]$: Time of spike at the $n^{th}$ layer.
    \item $W[n]$: Learnable weight at the $n^{th}$ layer.
\end{itemize}

The $\partial t[n] / \partial W[n]$ is calculated based on local learning rules of STDP and is equivalent to $\Delta w$ given in the Equation \ref{stdp_weight_updation}. SSTDP effectively combines the global optimization capabilities of backpropagation with the local weight adaptation of STDP, enabling SNNs to achieve competitive classification accuracy with significantly lower latency and computational cost.

\subsubsection{ANN to SNN Conversion}
ANN-to-SNN conversion is a method that not only completely circumvents the \textit{dead neuron} problem faced during backpropagation-based SNN training but also capitalizes on the existing repertoire of well-established ANN models. This approach converts a pretrained ANN model for a specific task into an SNN by replacing the activation function with LIF or IF. While the LIF activation function bears similarity to ReLU function when the spikes are accumulated over some time window \cite{diehl2015fast}, there have also been works \cite{wang2023new} that propose novel activations catering to the loss incurred during the ANN-to-SNN conversion. 
However, the resulting SNNs from ANN-to-SNN conversions are accompanied with some limitations \cite{deng2020rethinking}. Firstly, the ANN-to-SNN conversion method results in lossy conversion reducing the accuracy as compared to the original ANN. Secondly, the resulting SNN lacks any temporal dynamics that results in reduction of energy efficiency as it requires a longer time window to reach suitable accuracy as compared to directly trained SNNs. To address this issue, after an ANN is first converted to SNN, a fine-tune stage is introduced to learn the temporal relations.


\textbf{\textit{Remark: }}The training of SNNs presents unique challenges due to their event-driven and non-differentiable nature. While traditional gradient-based methods such as BPTT and STDP use clever continuous surrogate function as a substitute for backpropagation, methods like SpikeProp and SSTDP utilize time as a continuous variable for backpropagation, bypassing gradient-based approaches. Additionally, biologically-inspired approaches like STDP provide alternative mechanisms that emphasize efficiency and plausibility. For more detailed descriptions of SNN training strategies, please refer to~\cite{eshraghian2023training, Manon2024}.

%% file: sections/04_time_series_analysis.tex
\section{Time Series Analysis with SNN}
\label{sec:snn_applications}

In this section, we survey research that leverages SNNs for time series analysis in ubiquitous computing applications, spanning human activity recognition, speech classification, physiological signal processing, and emotion recognition. SNNs, with their event-driven and temporally precise processing, offer significant advantages for handling sequential sensor data, making them well-suited for real-time and energy-efficient computation. Our discussion covers key architectural choices, training methodologies, datasets, and the benefits of spike-based computation in these applications, highlighting the growing role of SNNs in pervasive and embedded intelligence.

\begin{figure}[ht]
    \centering
    \includegraphics[width=\linewidth]{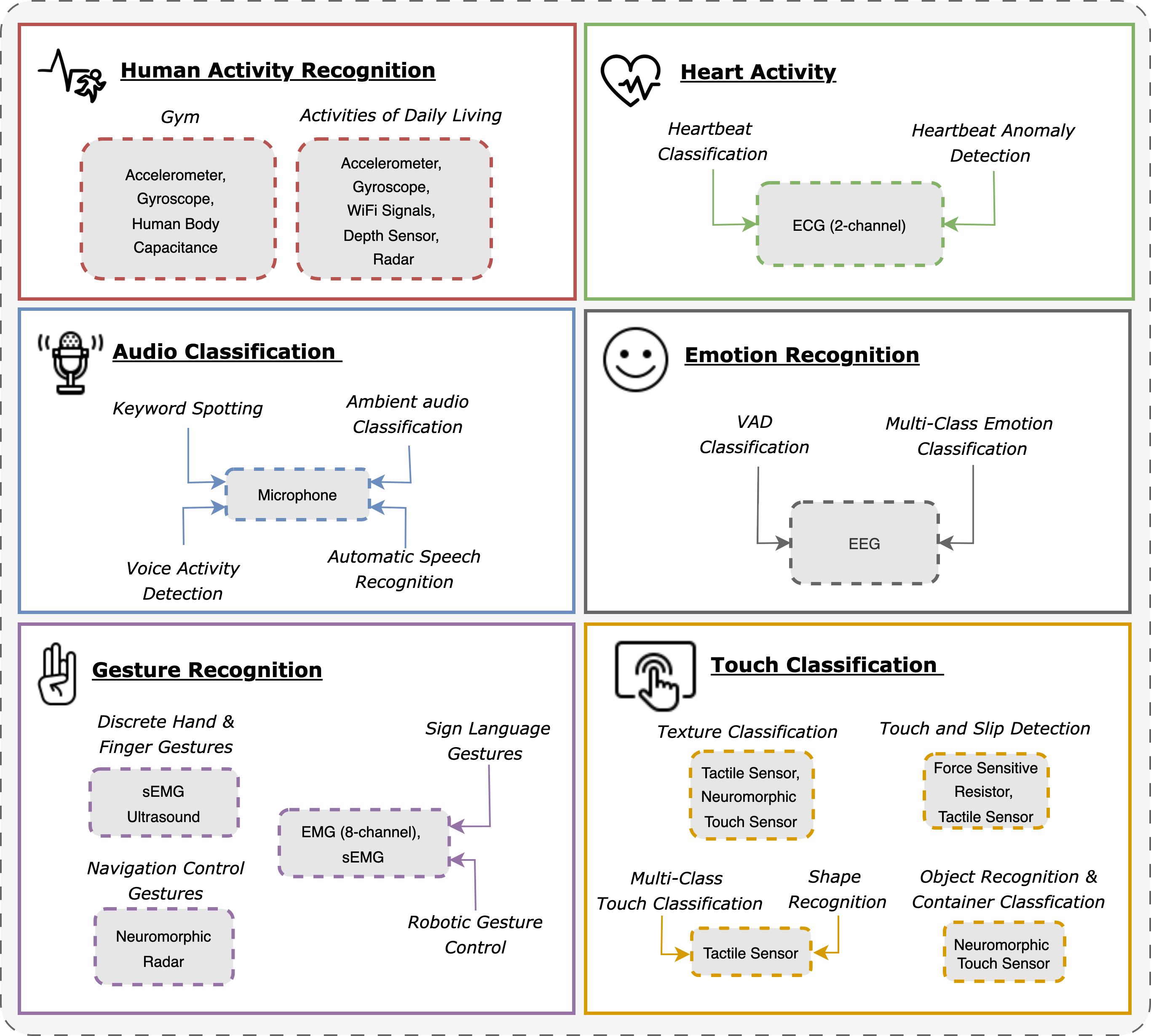} 
    \caption{Taxonomy based on the applications using SNNs.}
    \label{fig:taxonomy-applications}
\end{figure}


\subsection{Taxonomy}


We present a taxonomy that categorizes the use of SNNs for time series analysis in ubiquitous computing applications. This taxonomy organizes prior research into six key application domains, each encompassing specific use cases that leverage the event-driven and low-power properties of SNNs. As illustrated in Figure \ref{fig:taxonomy-applications}, these domains include Human Activity Recognition (HAR), Heart Activity Monitoring, Audio Classification, Emotion Recognition, Gesture Recognition, and Touch Classification. By structuring the field in this way, we highlight the diverse ways SNNs are applied to process continuous sensor data efficiently, enabling real-time and adaptive intelligence in wearable and pervasive systems.

The following subsections examine the prior work in each application domain, detailing the specific use cases, the goals they aim to achieve, and the sensor modalities leveraged to enable these applications.

\subsection{Human Activity Recognition (HAR)}
Human Activity Recognition (HAR) is a fundamental task in ubiquitous computing, enabling applications in healthcare monitoring, smart environments, and interactive systems~\cite{zhou2020deep, serpush2022wearable, irvine2019neural, singh2017convolutional}. However, real-world HAR presents several challenges, including the need for real-time processing~\cite{abdellatef2025detection}, energy-efficient computation~\cite{alsaadi2025logical}, and the ability to handle continuous, high-dimensional sensor data~\cite{sen2024continuous}. Traditional deep learning approaches, while effective, often rely on frame-based (i.e., at discrete time intervals) processing, leading to limitations such as redundant computations and inefficient power usage.

SNNs offer an alternative and efficient approach to HAR by leveraging their event-driven nature, temporal precision, and energy efficiency. Unlike conventional models that process data in fixed intervals, SNNs activate only when meaningful changes occur, significantly reducing computational overhead. Their inherent ability to capture fine-grained temporal dependencies makes them particularly suited for recognizing complex human activities while operating efficiently on neuromorphic hardware. By integrating SNNs, HAR systems can achieve lower latency, improved robustness to noisy sensor inputs, and enhanced adaptability, making them well-suited for real-world, always-on deployment.

In the literature, SNNs have primarily been applied to HAR across two main domains: (i) gym-based physical activities, which involve structured, high-intensity movements, and (ii) Activities of Daily Living (ADLs), characterized by more complex, subtle, and continuous movements that occur over extended durations. In the following section, we survey the research conducted in these two domains. 


\subsubsection{Gym Activities}
Prior research \cite{bian2024evaluation,10.1145/3594738.3611369} has primarily focused on the application of SNNs for gym activity recognition, utilizing both Inertial Measurement Unit (IMU) data and Human Body Capacitance (HBC) to classify 12 distinct exercises in the RecGym~\cite{bian2022contribution} dataset. These studies systematically explore various spiking encoding techniques for ubiquitous sensing, evaluating their efficacy using IMU + HBC data~\cite{bian2024evaluation}, with a three-layer SNN of 256, 64, and 12 neurons in each layer.
The encoding methods examined include rate encoding with diverse probability distributions (beta, uniform, normal), temporal encoding (linear, logarithmic), binary encoding (6-bit, 10-bit), and delta modulation with multiple threshold levels. Experimental results, trained via direct backpropagation, highlight that multi-threshold delta encoding (MT-delta) offers superior robustness compared to alternative schemes. This line of work was subsequently extended to develop an end-to-end neuromorphic system leveraging the MT-delta encoding approach, achieving the lowest energy-delay product (µJs) and benchmarking performance on neuromorphic hardware (Loihi) relative to general-purpose RISC-V and Cortex M7 processors \cite{10.1145/3594738.3611369}.


\subsubsection{Activities of Daily Living (ADLs)}
SNNs have also been explored for ADL recognition, particularly leveraging IMU data. Prior studies \cite{li, li2} employ SNNs to classify various activities (e.g., walking, jogging, sitting) using publicly available datasets such as UCI-HAR \cite{anguita2013public}, UniMB SHAR \cite{micucci2017unimib}, and HHAR \cite{stisen2015smart}. These works adopt an ANN-to-SNN conversion strategy, a commonly used technique that transforms a trained ANN into an SNN while preserving its learned representations. Energy consumption is assessed using the neuromorphic energy simulator Sata \cite{10.1109/TCAD.2022.3213211}, demonstrating a 94\% reduction in energy consumption compared to conventional ANNs while maintaining competitive classification performance.


Other studies \cite{suriani2021smartphone, fra2022human} investigate ADL classification using the WISDM dataset\cite{wisdm_smartphone_and_smartwatch_activity_and_biometrics_dataset__507}. Specifically, \cite{suriani2021smartphone} applies rate-based encoding with a supervised STDP training method, achieving 99.5\% accuracy across six smartphone-based activities (such as walking, eating, and drinking). In contrast, \cite{fra2022human} examines smartwatch-based activities (e.g., typing, writing, clapping) using spiking-equivalent models of CNNs and Legendre Memory Units (LMUs), typically with 2-3 Convolutional layers followed by 1-2 dense layers and comprising 90-160 neurons in it. 
They utilize LIF neurons for spike encoding, trained via ANN-to-SNN conversion, achieving 93.47\% and 94.51\% accuracy for seven activity classes, respectively. 


Beyond IMU-based sensing, a few studies have explored alternative sensory modalities such as Wi-Fi, depth sensing and radar for ADL recognition. In \cite{tan2021brain}, SNNs are used to analyze Wi-Fi signal reflections for human activity recognition, benchmarking their performance against LSTM, Bi-LSTM, and Gated Recurrent Unit (GRU) models. The results indicate that SNNs maintain competitive accuracy (84.8\%) compared to the highest-performing GRU model (88.57\%) while achieving over 70\% memory savings. Additionally, \cite{yang2018real, shen2021dynamic} utilize depth-sensing data from the MSR3D dataset for human motion detection, feeding time-series joint motion data into SNNs trained via BPTT. \cite{yang2018real} introduces a novel temporal coding scheme, achieving 75\% accuracy, while \cite{shen2021dynamic} enhances this to 83.6\% by incorporating recurrent SNNs with a simple LIF-based conversion. Furthermore, \cite{radar} used privacy-preserving radar sensing to classify 8 action classes from self-collected micro-Doppler data. They binarized the radar spectrograms to serve as spiking input for a convolutional SNN, achieving 85\% accuracy.

\subsubsection{\textbf{Key Takeaways}}
SNNs have been extensively studied in the context of HAR across various datasets utilizing diverse sensing modalities, including IMUs, Wi-Fi, depth sensors, and radar. As summarized in Table \ref{tab:HAR}, SNNs consistently achieve competitive performance, surpassing 83\% accuracy in most studies. Notably, on widely used public HAR benchmarks such as UCI-HAR and HHAR, state-of-the-art SNN-based models achieve accuracy exceeding 95\%. A recurring theme across these works is the substantial reduction in energy consumption and memory footprint offered by SNNs compared to conventional ANN architectures, while maintaining comparable classification performance.

Despite these promising results, further investigation is required to evaluate the feasibility of SNNs for continuous monitoring on ultra-edge devices. Prior works have already demonstrated the performance of SNNs on Loihi; however, assessing their deployment on low-power neuromorphic hardware platforms such as Xylo \cite{synsense_xylo}, T1 \cite{innatera_t1}, and DYNAP-SE2 \cite{richter2024dynap} is crucial for several reasons: (i) as we will introduce in Section~\ref{sec:hardware}, these platforms are specifically designed for energy-constrained edge applications, and benchmarking SNNs on them could reveal the true extent of power savings achievable in real-world settings, (ii) unlike Loihi, which remains a research-oriented neuromorphic processor, these emerging hardware platforms offer commercial viability for large-scale deployment, and (iii) exploring their efficiency in handling real-time, continuous HAR tasks will help determine their suitability for always-on wearable and embedded sensing applications, where ultra-low power consumption is a primary constraint.

\begin{table}[h!]
\centering
\resizebox{\textwidth}{!}{%
\begin{tabular}{|l|l|l|l|l|}
\hline
\rowcolor[HTML]{CBCBCB} 
\textbf{Paper} & \textbf{Spike Encoding} & \textbf{Training Method} & \textbf{Dataset} & \textbf{Accuracy} \\ \hline
\rowcolor[HTML]{F9F9F9} 
Bian et al. \cite{10.1145/3594738.3611369}  & MT-Delta & BPTT & RecGym \cite{bian2022contribution} & 87.5\% \\
\hline
\rowcolor[HTML]{FFFFFF} 
Bian et al. \cite{bian2024evaluation} & \begin{tabular}[c]{@{}l@{}} Rate \\ Temporal \\ Binary \\ MT-Delta\end{tabular} & BPTT & RecGym \cite{bian2022contribution} & \begin{tabular}[c]{@{}l@{}} 91.7\% \\ 89.1\%  \\ 89.6\% \\  89.8\% \end{tabular} \\
\hline

\rowcolor[HTML]{F9F9F9} 
Li et al. \cite{li}, \cite{li2} & LIF & ANN-to-SNNs & \begin{tabular}[c]{@{}l@{}}UCI-HAR \cite{anguita2013public}\\ UniMB SHAR \cite{micucci2017unimib}\\ HHAR \cite{stisen2015smart}\end{tabular} & \begin{tabular}[c]{@{}l@{}} 98.86\% \\ 94.04\% \\ 97.52\% \end{tabular}  \\
\hline

\rowcolor[HTML]{FFFFFF} 
Suriani et al. \cite{suriani2021smartphone}  & Rate & SSTDP & WISDM \cite{wisdm_smartphone_and_smartwatch_activity_and_biometrics_dataset__507} & 99.5\% \\
\hline

\rowcolor[HTML]{F9F9F9} 
Fra et al. \cite{fra2022human} & LIF & ANN-to-SNNs   & WISDM \cite{wisdm_smartphone_and_smartwatch_activity_and_biometrics_dataset__507} & \begin{tabular}[c]{@{}l@{}}  94.51 \end{tabular}\\
\hline

\rowcolor[HTML]{FFFFFF} 
Tan et al. \cite{tan2021brain}  & Temporal &  SSTDP & Kazuki  & 84.8\% \\
\hline
\rowcolor[HTML]{F9F9F9} 
Yang et al. \cite{yang2018real} &   Novel Temporal & BPTT  & MSR3D \cite{li2010action} & 75\% \begin{tabular}[c]{@{}l@{}} \end{tabular}\\ \hline
\rowcolor[HTML]{FFFFFF} 
Shen et al. \cite{shen2021dynamic} & LIF & BPTT & MSR3D \cite{li2010action} & 83.6\% \\
\hline
\rowcolor[HTML]{F9F9F9} 
Banerjee et al. \cite{radar} & Binary Spectrogram & STDP  & Radar actions (Self-collected) & 85\% \\
\hline
\end{tabular}%
}
\caption{Summary of SNN-based human activity recognition works. }
\label{tab:HAR}
\end{table}

\subsection{Heart Activity}

In the ubiquitous computing domain, detecting anomalies in heartbeat activity has become an active area of research, driven by the need for real-time, low-power, and always-on monitoring in wearable systems. These requirements demand models that can operate efficiently on resource-constrained edge devices without sacrificing accuracy. Among various physiological signals, electrocardiogram (ECG) signals have emerged as the preferred modality for non-invasive and continuous cardiac monitoring. 
Prior work in this space can be broadly categorized into two lines: (i) Heartbeat Classification~\cite{8879613, ecgIJCNN2019, yan2021energy, 9644939, feng2022building, chu2022neuromorphic, yan2024sparrowsnn, rana2024electrocardiography, banerjee2024atrial, banerjee2022snn}, 
and (ii) Heartbeat Anomaly Detection~\cite{8896021, 9948627, huang2024device, banerjee2024atrial, banerjee2022snn}. While heartbeat classification involves classification of heartbeat patterns such as Atrial Bigeminy, Atrial Fibrillation, Atrial Flutter, and Ventricular Bigeminy, among others, heartbeat anomaly detection is typically framed as a binary classification problem, wherein all abnormal heartbeat patterns are aggregated into a single "anomalous" class.

\subsubsection{Heartbeat Classification} A growing body of work has explored the use of deep SNNs for heartbeat classification, primarily via ANN-to-SNN conversion techniques. For instance, works such as~\cite{feng2022building} demonstrate that converting a CNN to its spiking equivalent can yield competitive performance, achieving 84.41\% accuracy on the PhysioNet 2017 dataset (exceeding the original ANN baseline by 0.35\% when simulated over extended time windows; ~1000 time steps). Similarly,~\cite{rana2024electrocardiography} integrate attention mechanisms into the ANN-SNN conversion pipeline, achieving 85.8\% on PhysioNet and 93.8\% accuracy on MIT-BIH Arrhythmia within only five-time steps and a modest power consumption of 278 mW.

Other studies such as~\cite{yan2021energy} adopt a modular two-stage CNN pipeline: first identifying abnormal heartbeats and then classifying the anomaly, which is subsequently converted to SNNs to reduce power demands, achieving 94.5\% accuracy on MIT-BIH while consuming 77 mW. Further advancements include the introduction of novel encoding schemes tailored to the spiking paradigm. For example, \cite{banerjee2022snn} propose peak encoding and compares it to encodings like gaussian encoder and delta modulator. They further propose a Spiking Hamilton encoding~\cite{banerjee2024atrial} and perform a systematic evaluation across multiple encoding methods (e.g., Gaussian, Delta, Traditional vs. Spiking Hamilton) and SNN architectures (reservoir and feedforward), reporting improved accuracy and robustness. Additionally, studies such as~\cite{ecgIJCNN2019} employ modified delta encoding in conjunction with reservoir-based SNNs to handle multi-class classification, achieving 95.6\% accuracy on the 17-class MIT-BIH Atrial Fibrillation dataset.

A few other works have also prioritized low-power implementations. For instance,~\cite{8879613} leverage unsupervised STDP learning followed by a reward-modulated STDP classifier, achieving competitive accuracy while drastically reducing energy consumption—down to 1.78 $\mu J$ per heartbeat, compared to 35 mJ for the best-performing ANN-based alternative. Custom hardware implementations such as those in~\cite{chu2022neuromorphic, 9644939} report 98.22\% accuracy on MIT-BIH with only 0.75 $\mu J$ per inference using ASIC-based deployment. More recent work such as~\cite{yan2024sparrowsnn} further improves energy efficiency, achieving 98.29\% accuracy at an ultra-low power budget of 0.031 $\mu J$ per classification.

\subsubsection{Heartbeat Anomaly Detection} 
For this task, a notable subset of this work leverages recurrent SNN architectures, particularly reservoir SNNs (rSNNs), which are well-suited for capturing temporal dependencies in physiological signals. For instance, works such as~\cite{8896021} implement an rSNN model on the Dynamic Neuromorphic Asynchronous Processor (DYNAP-SE) neuromorphic platform, achieving 91.3\% accuracy on the MIT-BIH Arrhythmia dataset while maintaining low power consumption at 722.1 $\mu W$. Similarly,~\cite{banerjee2024atrial} report 92\% accuracy on the PhysioNet 2017 dataset using a reservoir-based SNN model architecture.

Additional SNN model architectures have been proposed beyond reservoir models. For example,~\cite{9948627} introduce parallel delay chains to capture temporal patterns in ECG signals, also deploying the model on DYNAP-SE hardware. In another approach,~\cite{yan2021energy} utilizes a two-stage spiking CNN pipeline where the first stage performs anomaly detection before passing outputs to a classification module. Their model achieved 79\% anomaly detection accuracy on the MIT-BIH Arrhythmia dataset, consuming only 64 mW -- only 0.7\% of the power required by a comparable CNN. More recently,~\cite{huang2024device} improved on both accuracy and energy efficiency using a Spiking ConvLSTM 2D architecture, achieving 94.1\% accuracy for cardiac anomaly detection with energy consumption of 4.68 $\mu J$ per inference, almost a ten-fold reduction of 450 $\mu J$ reported for its ANN counterpart, with only a 0.7\% reduction in accuracy.

Spike encoding and training strategies have been found to significantly impact SNN performance in heartbeat anomaly detection. Supervised learning approaches frequently utilize delta coding~\cite{8896021,ecgIJCNN2019,9948627} or LC sampling~\cite{chu2022neuromorphic,9644939}. Among delta-coded models, rSNN architectures have shown higher accuracy—for example, 91.3\% and 95.6\% in~\cite{8896021} and~\cite{ecgIJCNN2019}, respectively. In contrast, LC sampling appears more effective for MLP-based SNNs, with accuracies reaching 95.34\% and 98.22\% in successive implementations by~\cite{chu2022neuromorphic,9644939}. For ANN-to-SNN converted models, Integrate-and-Fire (IF) encoding~\cite{rana2024electrocardiography,yan2024sparrowsnn} outperforms rate coding, achieving 93.8\% and 98.28\% accuracy compared to 90\% using rate coding~\cite{yan2021energy}. Unsupervised learning methods also show mixed results: while rate coding achieves better performance than Spiking Hamilton encoding in~\cite{8879613} and~\cite{banerjee2024atrial}, these outcomes may be architecture-dependent and not directly comparable. 

\subsubsection{\textbf{Key Takeaways}} Table \ref{tab:heart_activity} summarize existing SNN works for heart activity monitoring. The choice of spike encoding schemes and training methods plays a crucial role in determining the performance of SNNs for heart activity monitoring. Key aspects to consider include the effectiveness of delta coding in reservoir-based rSNNs and LC sampling in MLP-based architectures. Additionally, when transitioning from ANN to SNN models, the choice between IF encoding and rate coding can significantly impact model performance. Given the architectural dependencies, optimizing encoding and training strategies for specific SNN designs is essential.

From a practical standpoint, achieving real-time, energy-efficient systems remains a significant challenge. Hardware-software co-design offers a promising avenue to simultaneously enhance both accuracy and power efficiency in SNN-based solutions for heart activity monitoring. Future work should focus on advancing anomaly detection methods that minimize reliance on pre-segmented data, as well as refining SNN training processes to better exploit their temporal dynamics. These efforts will be pivotal in developing scalable, low-power, and personalized heart activity monitoring systems.

\begin{table}[h!]
\centering
\resizebox{\textwidth}{!}{%
\begin{tabular}{|l|l|l|l|l|}
\hline
\rowcolor[HTML]{CBCBCB} 
\textbf{Paper} & \textbf{Spike Encoding} & \textbf{Training Method} & \textbf{Dataset} & \textbf{Accuracy} \\
\hline

\rowcolor[HTML]{F9F9F9} 
Banerjee et al. \cite{banerjee2022snn} & \begin{tabular}[c]{@{}l@{}}Gaussian Encoding\\ Delta Encoding\\ Peak Encoding \end{tabular} & \begin{tabular}[c]{@{}l@{}}STDP \\ + Logistic Regression \end{tabular} & \begin{tabular}[c]{@{}l@{}}ECG200 \cite{olszewski2001generalized} \\ ECG5000 \cite{chenecg5000} \\ ECGFiveDays \cite{dau2019ucr} \\ PhysioNet/CinC 2017 \cite{clifford2017af} \\ MIT-BIH Arrhythmia \cite{moody2001impact} \end{tabular}  & \begin{tabular}[c]{@{}l@{}} 83\%  \\  93.1\%  \\ 88.9\%  \\ 78.5\%  \\ 94.3\% \end{tabular}  \\
\hline

\rowcolor[HTML]{FFFFFF} 
Banerjee et al. \cite{banerjee2024atrial} & Spiking Hamilton & \begin{tabular}[c]{@{}l@{}}STDP \\ + Logistic Regression \end{tabular} & \begin{tabular}[c]{@{}l@{}}ECG200 \cite{olszewski2001generalized} \\ ECG5000 \cite{chenecg5000} \\ ECGFiveDays \cite{dau2019ucr} \\ PhysioNet/CinC 2017 \cite{clifford2017af} \\ MIT-BIH Arrhythmia \cite{moody2001impact} \end{tabular}  & \begin{tabular}[c]{@{}l@{}} 82\%  \\  93.5\%  \\ 89.2\%  \\ 78.5\%  \\ 94.5\% \end{tabular}  \\
\hline

\rowcolor[HTML]{F9F9F9}
Rana et al. \cite{rana2024electrocardiography} & Integrate \& Fire & ANN-to-SNN &  \begin{tabular}[c]{@{}l@{}}MIT-BIH Arrhythmia \cite{moody2001impact} \\ PhysioNet/CinC 2017 \cite{clifford2017af} \end{tabular} & \begin{tabular}[c]{@{}l@{}} 93.8\% \\ 85.8\% \end{tabular}\\ 
\hline

\rowcolor[HTML]{FFFFFF}
Huang et al. \cite{huang2024device} & Rate & BPTT & TNMG \cite{ribeiro2020automatic} & 94.1\% \\
\hline

\rowcolor[HTML]{F9F9F9}
Yan et al. \cite{yan2024sparrowsnn} & Integrate \& Fire & ANN-to-SNN & MIT-BIH Arrhythmia \cite{moody2001impact} & 98.29\% \\
\hline

\rowcolor[HTML]{FFFFFF}
Chu et al. \cite{chu2022neuromorphic} & LC Sampling & STBP & MIT-BIH Arrhythmia \cite{moody2001impact} & 98.22\% \\
\hline 

\rowcolor[HTML]{F9F9F9}
Feng et al. \cite{feng2022building} & Rate  & ANN-to-SNN & PhysioNet/CinC 2017 \cite{clifford2017af} & 84.4\% \\
\hline

\rowcolor[HTML]{FFFFFF}
Gerber et al. \cite{9948627} & MT-Delta  & Parallel Delay Chains & \begin{tabular}[c]{@{}l@{}}PTB Diagnostic ECG \cite{bousseljot1995nutzung} \\ MIT-BIH Arrhythmia \cite{moody2001impact} \end{tabular} & \begin{tabular}[c]{@{}l@{}} 86.3\% \\ 82\%\end{tabular}  \\ 
\hline

\rowcolor[HTML]{F9F9F9}
Chu et al. \cite{9644939} & LC Sampling & STBP & MIT-BIH Arrhythmia \cite{moody2001impact} & 95.34\% \\
\hline

\rowcolor[HTML]{FFFFFF}
Yan et al. \cite{yan2021energy} & Rate & ANN-to-SNN & MIT-BIH Arrhythmia \cite{moody2001impact} & 90\% \\
\hline

\rowcolor[HTML]{F9F9F9} 
Corradi et al. \cite{ecgIJCNN2019} & Delta  & Supervised-R & MIT-BIH Atrial Fibrillation \cite{moody1983new} & 95.6\% \\
\hline

\rowcolor[HTML]{FFFFFF}
Amirshahi et al. \cite{8879613} & Rate  & STDP and RSTDP & MIT-BIH Arrhythmia \cite{moody2001impact} & 97.9\% \\
\hline

\rowcolor[HTML]{F9F9F9}
Bauer et al. \cite{8896021} & Delta (Sigma-delta) & Supervised-R & MIT-BIH Arrhythmia \cite{moody2001impact} & 91.3\% \\
\hline

\end{tabular}%
}
\caption{Summary of SNN-based heart activity works. }
\label{tab:heart_activity}
\end{table}


\subsection{Audio Classification}

The ability of SNNs to efficiently process temporal signals makes them particularly well-suited for audio signal processing tasks. 
In the domain of audio processing, research using SNNs has primarily focused on four key areas: (i) Keyword Spotting (KWS), (ii) Automatic Speech Recognition (ASR), (iii) Voice Activity Detection (VAD), and (iv) Ambient Audio Classification (AAC).


It is important to distinguish between the tasks of KWS and ASR, as they involve different classification paradigms. ASR is inherently a multi-class classification problem, where the model is tasked with identifying and classifying spoken words from a predefined lexicon. In contrast, KWS is framed as a one-vs-all binary classification problem, where the model classifies spoken input as either a positive (keyword) or negative (non-keyword) detection. VAD also represents a binary classification task, but instead of identifying specific words, the model determines the presence or absence of speech within the audio signal. AAC, on the other hand, is a multi-class classification task where the model categorizes ambient sounds into different environmental contexts, such as "cafe", "home", "car", or "street". Developing low-power solutions for AAC is particularly relevant for real-time deployment in energy-constrained devices such as headsets, smartphones, and hearing aids, where efficient ambient noise filtering can significantly enhance user experience by reducing background noise interference.

\subsubsection{Keyword Spotting (KWS)}

Recent work has shown that SNNs can achieve competitive performance on KWS while significantly reducing energy consumption. For instance, Blouw et al.~\cite{10.1145/3320288.3320304} trained a feed-forward SNN on the Aloha dataset~\cite{10.1145/3320288.3320304}, which consists of utterances of the keyword "Aloha" interleaved with distractor phrases. Their model achieved 93.8\% accuracy and demonstrated a 110$\times$ reduction in energy per inference when deployed on Intel's Loihi chip (0.27 $mJ$) compared to a GPU (29.8 $mJ$). Building on this, Bos et al.~\cite{bos2024micro} proposed SynNet, a fully connected architecture with LIF neurons implemented on the Xylo neuromorphic hardware. They attained a higher accuracy of 95.31\% and further reduced dynamic energy consumption to just 6.6$\mu J $ per inference.

While these approaches used standard preprocessing pipelines, other efforts have explored more biologically inspired spike encodings to better exploit the SNN architecture. Cramer et al.\cite{cramer2020heidelberg} addressed the lack of standardized input formats by introducing the Spiking Speech Commands (SSC) and Spiking Heidelberg Digits (SHD) datasets. Using a human auditory-inspired encoding (i.e., simulating cochlear frequency decomposition and spiking via hair and bushy cells), they transformed the Speech Commands~\cite{warden1804speech} and newly recorded spoken digits into realistic spike trains. Their SNN model operating in a reservoir paradigm achieved 83.2\% accuracy on SSC and 50.9\% on SHD. Perez-Nieves et al.~\cite{perez2021neural} extended this by introducing heterogeneity in neuron membrane time constants, reaching 81.7\% and 56.1\% on the same datasets, respectively.

To move beyond reliance on handcrafted or auditory-inspired encodings, some works have proposed end-to-end architectures that remove explicit preprocessing. Auge et al.\cite{auge2021end}, for example, used resonate-and-fire input neurons that directly respond to audio frequency components, eliminating the need for feature extraction or buffering. Similarly, Weidel et al.~\cite{weidel2021wavesenseefficienttemporalconvolutions} introduced WaveSense, an SNN model inspired by the WaveNet architecture~\cite{van2016wavenet}, capable of real-time inference without any audio buffering. Their model achieved state-of-the-art results on several KWS benchmarks: 99.5\% on Aloha~\cite{10.1145/3320288.3320304}, 99.8\% on Hey Snips~\cite{coucke2019efficient}, and 87.6\% on Speech Commands~\cite{warden1804speech}. Complementary studies such as Yilmaz et al.~\cite{yilmaz2020deep} also offered comparative insights into ANN and SNN versions of DNNs and CNNs, highlighting trade-offs in accuracy versus energy.

\subsubsection{Automatic Speech Recognition (ASR)}
SNN models for ASR typically fall into three broad categories: reservoir-based SNNs, spiking CNNs, and feed-forward MLP-based SNNs using IF or LIF neurons. These architectures have shown promise because they align well with the temporal and spectral nature of speech: (a) reservoir SNNs leverage recurrent spiking dynamics to retain temporal context, making them effective for modeling the sequential nature of speech, (b) spiking CNNs extract localized spectro-temporal features using convolutional layers, enabling accurate recognition of phonemes and frequency patterns in structured auditory inputs, and (c) MLP-based SNNs offer energy efficiency, relying on effective spike encodings and lightweight architectures for real-time ASR in constrained environments.

Reservoir-based SNNs leverage dynamic temporal patterns in spiking activity for processing sequential data. Cramer et al.\cite{cramer2020heidelberg} demonstrated this using their Spiking Heidelberg Digits dataset, achieving an accuracy of 83.2\%. Building on this, Perez-Nieves et al.\cite{perez2021neural} introduced synaptic and membrane heterogeneity in the reservoir, resulting in comparable accuracy (81.7\%) but improved robustness. While effective, these models tend to underperform compared to convolutional architectures.

Spiking CNNs (SCNN) have emerged as a more accurate alternative. Morales et al.\cite{dominguez2018deep} implemented a biologically inspired audio encoding pipeline using a Neuromorphic Auditory Sensor (NAS), followed by a spiking CNN on FPGA, achieving 89.9\% on a subset of the Speech Commands dataset. Other works, such as Dong et al.\cite{dong2018unsupervised}, employed unsupervised STDP to train spiking CNNs followed by supervised classifiers like SVM or Tempotron, reporting accuracies of 97.5\% and 93.8\% on TIDIGITS~\cite{leonard1993tidigits} and TIMIT~\cite{garofolo1993timit}, respectively. Leow et al.~\cite{leow2023sparsity} further optimized SCNNs using Bayesian search for neuron parameters, achieving 98.4\% on the Free Spoken Digits Dataset (FSDD)~\cite{jackson2018jakobovski}.

Feed-forward SNNs using MLPs have also been explored. Dennis et al.\cite{dennis2015combining} proposed a novel spectrogram-based spike encoding and used a modified Tempotron rule, achieving 94\% accuracy on the RWCP dataset~\cite{nakamura2000acoustical}. More recently, Wu et al.\cite{wu2020deep} introduced a Self-Organizing Map-based representation and a Maximum-Margin Tempotron rule, achieving state-of-the-art performance of 99.6\% on RWCP and 97.6\% on TIDIGITS. Complementing this, Bensimon et al.~\cite{bensimon2021using} proposed an energy-efficient Spiking Continuous Time Neuron (SCTN) encoder with direct hardware implementation, bypassing the need for ADCs.

Hybrid training strategies like ANN-SNN tandem learning have also shown promise. Wu et al.~\cite{wu2020deep} used this approach to train deep SNNs with shared weights, achieving competitive error rates: 18.7\% phone error rate (PER) on TIMIT, and word error rates (WER) of 36.9\% and 9.4\% on the FAME and Librispeech datasets, respectively.

\subsubsection{Ambient Audio Classification (AAC)}
In AAC, several approaches have focused on energy-efficient classification of audio scenes, though there is a lack of a standardized benchmark dataset for consistent evaluation. For example, Bos et al.\cite{9967462} aimed at reducing power consumption for classifying ambient noise scenes, achieving 98.0\% accuracy with a total inference power consumption of 542 $\mu W$ on the Xylo platform. Similarly, Ke et al.\cite{ke2024neurobench} worked on acoustic scene classification, achieving 93\% accuracy on the DCASE 2020 dataset~\cite{mesaros2018multi} with a power consumption of 692 $\mu W$ and an energy usage of 57.6 $\mu J$ per inference on Xylo. In a different direction, Mukhopadhyay et al.\cite{mukhopadhyay2022acoustic} designed a low-power sensor node for acoustic scene surveillance, proposing a power-efficient feature extraction method that achieved an energy consumption of 6.8 $nJ$ per inference on the Xilinx Zynq-7000 processor. Lastly, Kshirasagar et al.\cite{kshirasagar2024auditory} worked on siren detection in audio scenes, developing both a conventional RNN and its equivalent Spiking RNN variant, which demonstrated 3x fewer parameters while maintaining over 90\% accuracy.


\subsubsection{ Voice Activity Detection (VAD)}
A variety of spike encoding schemes have been proposed to improve efficiency and performance. For example, Martinelli et al.\cite{martinelli2020spiking} utilized simple temporal coding, while Dellaferrera et al.\cite{dellaferrera2020bin} introduced bin-based encoding. More recently, Yang et al.~\cite{yang2024svad} integrated a spiking CNN version of SincNet with an attention mechanism, demonstrating substantial gains in energy efficiency, with power consumption in the micro-watt range.

Compared to traditional ANN-based VAD systems~\cite{meoni2018low, price2017low, yang20181muw}, which span from micro-watt to milli-watt power budgets, SNN-based approaches consistently deliver lower power footprints without significant trade-offs in accuracy. Wu et al.\cite{wu2021hurai} reported a 21.54$\times$ reduction in synaptic operations compared to an equivalent ANN model, with only a modest 1.44\% increase in Half Total Error Rate (HTER). Notably, Yang et al.\cite{yang2024svad} achieved state-of-the-art energy efficiency, operating at just 0.9~$\mu W$ while maintaining HTERs of 5.8\%, 12.6\%, and 22.3\% across low, medium, and high noise conditions, respectively. These developments highlight the promise of SNNs for always-on edge audio sensing with tight power constraints.

\begin{table}[h!]
\centering
\resizebox{\textwidth}{!}{%
\begin{tabular}{|l|l|l|l|l|l|}
\hline
\rowcolor[HTML]{CBCBCB} 
\textbf{Paper} & \textbf{Application} & \textbf{Spike Encoding} & \textbf{Training Method} & \textbf{Dataset} & \textbf{Accuracy} \\
\hline

\rowcolor[HTML]{F9F9F9} 
Blouw et al. \cite{10.1145/3320288.3320304} & Keyword Spotting & LIF & ANN-to-SNN & Aloha \cite{10.1145/3320288.3320304} & 93.8\%\\
\hline

\rowcolor[HTML]{FFFFFF}
Auge et al. \cite{auge2021end} & Keyword Spotting & Resonating Input Neurons & BPTT & Speech Commands \cite{warden1804speech}  & 86.70\%\\
\hline

\rowcolor[HTML]{F9F9F9} 
Weidel et al. \cite{weidel2021wavesenseefficienttemporalconvolutions} & Keyword Spotting & LIF & BPTT & \begin{tabular}[c]{@{}l@{}}Aloha \cite{10.1145/3320288.3320304}\\SSC \cite{cramer2020heidelberg} \\Hey Snips \cite{coucke2019efficient} \end{tabular} & \begin{tabular}[c]{@{}l@{}}98\%\\79.6\%\\99.6\%\end{tabular} \\ 
\hline

\rowcolor[HTML]{FFFFFF} 
Bos et al. \cite{bos2024micro} & Keyword Spotting & Power Band Bin to Spike & BPTT & Aloha \cite{10.1145/3320288.3320304} & 95.31\% \\ 
\hline

\rowcolor[HTML]{F9F9F9} 
Perez-Nieves et al. \cite{perez2021neural} & \begin{tabular}[c]{@{}l@{}}Keyword Spotting \\ Speech Recognition \end{tabular} & N/A & BPTT & \begin{tabular}[c]{@{}l@{}}SHD \cite{cramer2020heidelberg} \\ SSC \cite{cramer2020heidelberg} \end{tabular} & \begin{tabular}[c]{@{}l@{}} 81.7\%\\ 56.1\%\end{tabular} \\
\hline

\rowcolor[HTML]{FFFFFF} 
Cramer et al. \cite{cramer2020heidelberg} & \begin{tabular}[c]{@{}l@{}}Keyword Spotting \\ Speech Recognition \end{tabular} & N/A & BPTT & \begin{tabular}[c]{@{}l@{}}SHD \cite{cramer2020heidelberg} \\ SSC \cite{cramer2020heidelberg}\end{tabular} & \begin{tabular}[c]{@{}l@{}}83.2\%\\ 50.9\%\end{tabular} \\
\hline

\rowcolor[HTML]{F9F9F9}
Dennis et al. \cite{dennis2015combining} & Speech Recognition & Spectogram Features & Tempotron & RWCP \cite{nakamura2000acoustical} & 94.10\%\\
\hline

\rowcolor[HTML]{FFFFFF}
Dong et al. \cite{dong2018unsupervised} & Speech Recognition & Temporal Coding & STDP & \begin{tabular}[c]{@{}l@{}}TIDIGITS \cite{leonard1993tidigits} \\TIMIT \cite{garofolo1993timit} \end{tabular} & \begin{tabular}[c]{@{}l@{}}97.5\%\\3.8\%\end{tabular} \\
\hline

\rowcolor[HTML]{F9F9F9}
Dominguez et al. \cite{dominguez2018deep} & Speech Recognition & Cochlear Encoding & BPTT & Speech Commands \cite{warden1804speech} & 89.90\% \\
\hline

\rowcolor[HTML]{FFFFFF} 
Wu et al. \cite{wu2018biologically} & Speech Recognition & Spikes from SOM & Tempotron & TIDIGITS \cite{leonard1993tidigits} & 97.6\% \\
\hline

\rowcolor[HTML]{F9F9F9} 
Wu et al. \cite{audio3} & Speech Recognition & Spikes from SOM & Tempotron & \begin{tabular}[c]{@{}l@{}}RWCP \cite{nakamura2000acoustical} \\TIDIGITS \cite{leonard1993tidigits} \end{tabular} & \begin{tabular}[c]{@{}l@{}}99.6\%\\97.4\%\end{tabular} \\
\hline
\rowcolor[HTML]{FFFFFF}
Wu et al. \cite{wu2020deep} & Speech Recognition & IF & Tandem learning & \begin{tabular}[c]{@{}l@{}}TIMIT \cite{garofolo1993timit} \\FAME \cite{yilmaz2016longitudinal} \\Librispeech \cite{panayotov2015librispeech} \end{tabular} &\begin{tabular}[c]{@{}l@{}}PER: 18.7\%\\WER: 36.9\%\\WER: 9.4\%\end{tabular} \\
\hline

\rowcolor[HTML]{F9F9F9}
Bensimon et al. \cite{bensimon2021using} & Speech Recognition & SCTN based Resonator & STDP & RWCP \cite{nakamura2000acoustical} &  98.73\%\\
\hline

\rowcolor[HTML]{FFFFFF}
Leow et al. \cite{leow2023sparsity} & Speech Recognition & LIF & BPTT & FSDD \cite{jackson2018jakobovski} &  98.4\%\\
\hline

\rowcolor[HTML]{F9F9F9} 
Bos et al. \cite{9967462} & Ambient Audio Classification & Power Band bin to Spike & BPTT & QUT-NOISE \cite{dean2010qut} & 98.0\% \\
\hline

\rowcolor[HTML]{FFFFFF} 
Mukhopadhyay et al. \cite{mukhopadhyay2022acoustic} & Ambient Audio Classification & Rate Coding & ANN-to-SNN & Self-Collected \cite{mukhopadhyay2022acoustic} & 93.44\% \\
\hline

\rowcolor[HTML]{F9F9F9} 
Ke et al. \cite{ke2024neurobench} & Ambient Audio Classification & Power Band Bin to Spike & BPTT & DCASE 2020 \cite{mesaros2018multi} &  93.0\% \\
\hline

\rowcolor[HTML]{FFFFFF} 
Kshirasagar et al. \cite{kshirasagar2024auditory} & Ambient Audio Classification & N/A & BPTT & Public Dataset \cite{asif2022large} &  90.1\% \\
\hline

\rowcolor[HTML]{F9F9F9} 
Dellaferrera et al. \cite{dellaferrera2020bin} & Voice Activity Detection & Bin Encoding & Tempotron & QUT-NOISE TIMIT \cite{dean2010qut} & HTER: 8.2\% \\
\hline

\rowcolor[HTML]{FFFFFF}
Martinelli et al. \cite{martinelli2020spiking} & Voice Activity Detection & Temporal & BPTT & QUT-NOISE-TIMIT \cite{dean2010qut} & HTER: 4.6\%\\
\hline

\rowcolor[HTML]{F9F9F9} 
Wu et al. \cite{wu2021hurai} & Voice Activity Detection & LIF & BPTT & QUT-NOISE-TIMIT \cite{dean2010qut} & HTER: 2.72\% \\
\hline

\rowcolor[HTML]{FFFFFF}
Yang et al. \cite{yang2024svad} & Voice Activity Detection & SincNet + Attention & BPTT & QUT-NOISE TIMIT \cite{dean2010qut} & HTER: 4.0\%\\ 
\hline

\end{tabular}%
}
\caption{Summary of SNN-based audio classification works.}
\label{tab:audio_classification}
\end{table}

\subsubsection{\textbf{Key Takeaways}}
Table~\ref{tab:audio_classification} summarizes existing SNN-based audio classification works.
\begin{itemize}[leftmargin=*]
    \item KWS: Current SNN-based KWS models work well for fixed vocabularies in controlled conditions but struggle with variability in noise, accents, or speakers. In ubiquitous scenarios, KWS systems need to adapt on-device, learning new keywords, handling overlapping speech, and adjusting to new users. Techniques like few-shot learning, spike-based attention, and class-incremental learning are promising directions.

    \item ASR: SNNs offer low-power ASR, but most models rely on fixed feature encoders and shallow architectures, limiting performance on continuous speech or diverse speakers. For real-world UbiComp settings like smart homes or wearables, there’s a need for deeper, more adaptable spiking architectures that handle long-term context and noisy input. End-to-end spike-based encoders offer exciting potential to reduce latency and improve robustness at the edge.

    \item Research in ambient sound sensing with SNNs often relies on tailored datasets and task-specific models. There is a significant opportunity to establish more generalizable use cases, such as monitoring domestic activities or detecting anomalies in smart environments, while also creating standardized benchmarks for evaluation. Future work should focus on developing on-chip, spike-based encoders for features like MFCCs or spectrograms, coupled with few-shot or contrastive self-supervised learning techniques to enhance the system's ability to generalize to novel acoustic contexts with minimal retraining.

    \item VAD: Given the always-on nature of VAD applications, recent SNN-based approaches have prioritized energy efficiency while maintaining detection accuracy. Techniques like temporal and bin encoding, attention-guided SincNet architectures, and hybrid training strategies have led to $\mu W$-level implementations on edge devices. However, current systems are mostly evaluated on synthetic or noise-augmented datasets, which limits generalizability. Opportunities remain in designing spike encodings that adapt dynamically to background noise profiles, implementing online learning mechanisms for personalized noise conditions, and incorporating spike-based recurrent memory modules to better capture speech onset and offset in fluctuating environments.
\end{itemize}

\subsection{Emotion Classification}

Recent advances in this field have focused on multi-modal approaches combining EEG with other physiological signals~\cite{huang2017fusion, gohumpu2023emotion, lee2024encoding} and specialized feature extraction techniques~\cite{akhandimproved}. SNN-based emotion recognition typically addresses two classification paradigms: (i) dimensional (valence (V), arousal (A), dominance (D), and like/dislike (L)) using datasets such as DEAP \cite{koelstra2011deap}, DREAMER \cite{katsigiannis2017dreamer}, and  MAHNOB-HCI \cite{soleymani2011multimodal}, and (ii) discrete emotion classification using datasets such as SEED \cite{zheng2015investigating}, SEED-IV \cite{zheng2018emotionmeter}, and MPED \cite{song2019mped}.

Early work leverages the NeuCube architecture~\cite{kasabov2014neucube}, a recurrent SNN framework designed for spatio-spectro-temporal brain data (STBD). NeuCube comprises an encoder, a 3D spatially organized SNN reservoir (SNNr) trained via Spike-Timing Dependent Plasticity (STDP), and an evolving SNN classifier (eSNN). Luo et al.~\cite{luo2020eeg} explored a range of time-, frequency-, and time-frequency domain transformations such as variance, FFT, and DFT for encoding EEG input and achieved accuracies in the range of  74-86\% on dimensional classification, on DEAP, and 96.67\% on SEED. Alnafjan et al.\cite{al2020lightweight} focused on lightweight NeuCube variant implementations suitable for resource-constrained environments. However, these approaches often require significant preprocessing and struggle with subject-independent generalization. To address subject-dependence in EEG emotion labeling, Tan et al.\cite{tan2021neurosense} used facial landmark tracking to segment EEG into emotionally salient windows, achieving up to 78.97\% (A) on DEAP and MAHNOB-HCI. Alzahrani et al.\cite{alzhrani2021emotion} analyzed NeuCube’s evolving connectivity patterns and reported 83.5\% (binary) and 94.83\% (four-class) accuracy on the DREAMER dataset.

Moving beyond NeuCube, recent efforts have introduced new SNN architectures and training paradigms to improve classification performance and resource efficiency. Li et al.~\cite{li2023fractal} introduced FractalSNN, a novel architecture designed to capture multiscale temporal-spectral-spatial (TSS) dynamics in EEG signals for emotion classification. The core idea is to use parallel sub-networks of varying path lengths to extract features at different resolutions, whose outputs are fused to enrich the spatiotemporal representation. To improve robustness and prevent overfitting, they also proposed an inverted drop-path regularization scheme that randomly disables entire sub-paths during training. FractalSNN achieved notable performance, with up to 80.92\% (D) accuracy on DREAMER, and showed strong generalization across DEAP (73.20\%), SEED (68.33\%), and MPED (42.23\%), highlighting its cross-dataset adaptability.

Focusing on energy efficiency, Yan et al.~\cite{yan2022eeg} proposed an energy-efficient ANN-to-SNN conversion method, transforming a pretrained CNN into an SNN via a clamped and quantized transfer scheme. Their model preserved 91.43\% accuracy on the SEED dataset while reducing energy consumption to just 13.8\% of the original CNN. Building on this, Xu et al.~\cite{xu2024eescn} introduced Emo-EEGSpikeConvNet (EESCN), which integrates a neuromorphic data generation module (using differential entropy across EEG bands), a spiking encoder, and a spiking convolutional classifier. EESCN delivered 1.9–3.3$\times$ speedups over Yan et al.~\cite{yan2022eeg} and 6.4–7.9$\times$ over NeuCube~\cite{luo2020eeg}, while achieving comparable or better accuracy on SEED and DREAMER.

\textbf{\textit{Key Takeaways: }}
These efforts reflect a trajectory from biologically inspired but general-purpose SNN models to efficient, task-adaptive architectures that make better use of the structure inherent in EEG signals. Systems such as NeuCube, which utilize STDP and dynamic spatio-temporal state representations, are well-suited to capture the transient, distributed dynamics of emotional states in EEG signals. Recent architectures like FractalSNN and EESCN have shown improved performance and generalization across multiple datasets, moving closer to practical emotion recognition systems that can function across users and contexts. However, subject-specific variability continues to pose a significant challenge. Strategies such as facial landmark-based segmentation have shown promise in reducing this dependency, but more robust and generalizable approaches are still needed.

Despite these advancements, the field remains heavily reliant on a narrow set of laboratory datasets (e.g., DEAP, DREAMER), which limits the ability to assess model robustness under the noisy, variable conditions typical of everyday environments. From a Ubicomp perspective, this represents a critical gap and an opportunity: there is a pressing need for emotion recognition systems that are both ecologically valid and computationally efficient. SNNs, especially those optimized for low-power deployment through ANN-to-SNN conversion or custom spike encoding pipelines, hold significant promise for enabling always-on, on-device emotion sensing in wearables or mobile systems. Looking ahead, the integration of EEG with other modalities, such as facial expressions, galvanic skin response, or voice, remains an underexplored but promising direction.


\begin{table}[h!]
\centering
\resizebox{\textwidth}{!}{%
\begin{tabular}{|l|l|l|l|l|}
\hline
\rowcolor[HTML]{CBCBCB} 
\textbf{Paper} & \textbf{Spike Encoding} & \textbf{Training Method} & \textbf{Dataset} & \textbf{Accuracy} \\ \hline

\rowcolor[HTML]{F9F9F9} 
Xu et al. \cite{xu2024eescn} & LIF & BPTT & \begin{tabular}[c]{@{}l@{}}DEAP \cite{koelstra2011deap} \\ SEED-IV \cite{zheng2018emotionmeter} \end{tabular} & \begin{tabular}[c]{@{}l@{}}94.56\% (V), 94.81\% (A), 94.73\% (D) \\ 79.65\% (4-class)\end{tabular} \\ \hline

\rowcolor[HTML]{FFFFFF} 
Li et al. \cite{li2023fractal} & \begin{tabular}[c]{@{}l@{}}Preprocessed EEG \\ Temporal \\ PSD \\ DE\end{tabular} & BPTT & \begin{tabular}[c]{@{}l@{}}DREAMER \cite{katsigiannis2017dreamer} \\ DEAP \cite{koelstra2011deap} \\ SEED-IV \cite{zheng2018emotionmeter} \\ MPED \cite{song2019mped} \end{tabular} & \begin{tabular}[c]{@{}l@{}}71.01\% (V), 78.50\% (A), 80.92\% (D) \\ 69.84\% (V), 69.61\% (A), 73.20\% (D) \\ 68.33\% (4-class) \\ 42.23\% (7-class)\end{tabular} \\ \hline

\rowcolor[HTML]{F9F9F9} 
Yan et al. \cite{yan2022eeg} & IF & ANN-to-SNN & DEAP \cite{koelstra2011deap} & 82.75\% (V), 84.22\% (A) \\ \hline

\rowcolor[HTML]{FFFFFF} 
Alzhrani et al. \cite{alzhrani2021emotion} & TBR & \begin{tabular}[c]{@{}l@{}}STDP \\ + Supervised SDSP\end{tabular} & DREAMER \cite{katsigiannis2017dreamer} & \begin{tabular}[c]{@{}l@{}}83.5\% (2-class) \\ 94.83\% (4-class)\end{tabular} \\ \hline

\rowcolor[HTML]{F9F9F9} 
Tan et al. \cite{tan2021neurosense} & TBR & \begin{tabular}[c]{@{}l@{}}STDP \\ + Supervised SDSP\end{tabular} & \begin{tabular}[c]{@{}l@{}}DEAP \cite{koelstra2011deap} \\ MAHNOB-HCI \cite{soleymani2011multimodal} \end{tabular} & \begin{tabular}[c]{@{}l@{}}67.76\% (V), 78.97\% (A) \\ 78.97\% (A), 67.76\% (V)\end{tabular} \\ \hline

\rowcolor[HTML]{FFFFFF} 
Luo et al. \cite{luo2020eeg} & Temporal & \begin{tabular}[c]{@{}l@{}}STDP \\ + Supervised SDSP\end{tabular} & \begin{tabular}[c]{@{}l@{}}DEAP \cite{koelstra2011deap} \\ SEED \cite{zheng2015investigating} \end{tabular} & \begin{tabular}[c]{@{}l@{}}78\% (V), 74\% (A), 80\% (D), 86.27\% (L) \\ 96.67\% (3-class)\end{tabular} \\ \hline

\rowcolor[HTML]{F9F9F9} 
Alnafjan et al. \cite{al2020lightweight} & TBR & \begin{tabular}[c]{@{}l@{}}STDP \\ + Supervised SDSP\end{tabular} & DEAP \cite{koelstra2011deap} & 84.62\% (V), 61.54\% (A) \\ \hline
\end{tabular}%
}
\caption{Summary of SNN-based emotion recognition works.}
\label{tab:emotion_recognition}
\end{table}

\subsection{Gesture Recognition}

Human body gestures are a key modality for natural and intuitive interaction in wearables, smart environments, and gaming systems. For instance, contemporary smartwatches now incorporate fine-grained gestures such as pinch and double-pinch for interface control~\cite{xu2022enabling}, while ambient sensor networks enable gesture-driven actuation in smart homes~\cite{li2022trajectory}. 
SNNs offer a compelling alternative to conventional deep learning models by utilizing event-driven spike-based representations, where computation is triggered only in response to salient input changes. This inherent sparsity significantly reduces both energy consumption and computational overhead, making SNNs well-suited for deployment on low-power edge devices. Moreover, the temporal processing capabilities of SNNs naturally capture dynamic gesture patterns without the need for additional components such as LSTM layers or recurrent connections, thereby simplifying model architectures.

In this section, we organize prior work on gesture classification using SNNs into four distinct categories based on application contexts: (i) discrete hand and finger gestures, (ii) robotic gesture control, (iii) sign language recognition, and (iv) navigation control gestures. This taxonomy highlights the diversity of tasks addressed in the literature and allows for a more structured comparison of approaches and design choices.

\subsubsection{Discrete Hand and Finger Gestures}
A growing body of work has explored SNNs for surface electromyography (sEMG)-based gesture recognition, with a focus on deploying models on energy-efficient neuromorphic hardware. While many of these studies demonstrate the feasibility of using SNNs for real-time, low-power applications, they often trade off some classification accuracy compared to conventional deep learning models.

Several works have investigated direct deployment of SNNs on neuromorphic chips such as  DYNAP-SE and Loihi for hand gesture recognition. Donati et al.~\cite{8747378} and Ma et al.~\cite{9073810} both used the Myo armband to classify simple gestures like rock, paper, and scissors. Donati’s implementation on the DYNAP chip achieved modest accuracy relative to traditional classifiers but demonstrated extremely low power consumption. \cite{9073810} extended this by deploying a spiking recurrent neural network (SRNN) with delta-based spike encoding, showing that reservoir-based SNNs can outperform traditional SVM baselines when dynamic temporal features are preserved.

Meanwhile, Vitale et al.~\cite{9849452}, Zhang et al.~\cite{10.1016/j.ins.2021.11.065}, and Guo et al.~\cite{guo2024spgesture} each focused on classifying daily-life hand and finger gestures. Both  ~\cite{9849452, 10.1016/j.ins.2021.11.065} employed the NinaPro DB datasets: Vitale et al. proposed a spiking CNN–based network, while Zhang et al. introduced a novel second‑order information bottleneck (2O‑IB) loss function before converting their model into a spike‑based SNN. Although both approaches achieved moderate accuracy, Vitale et al. demonstrated their implementation on the Loihi chipset, achieving low latency (5.7ms) and power consumption of just 41mW. Efforts to improve both performance and generalizability are exemplified by Guo et al.’s SpGesture framework~\cite{guo2024spgesture}, which introduces a Spiking Jaccard Attention mechanism and source‑free domain adaptation to handle variability across different forearm postures. Using a ConvLIF encoder and SuperSpike training, their model delivers high accuracy while preserving low latency, making it well‑suited for real‑world wearable applications where signal variability is common.

Additionally, ~\cite{echoWrite, stereoGest} have used ultrasound as a sensory modality for gesture recognition. Arun et al.~\cite{echoWrite} employed a 5-layer CNN model converted into an SNN to classify air-written finger gestures, achieving 92\% accuracy and a $4.4\times$ reduction in computational cost compared to the original CNN. In contrast, Andrew et al.~\cite{stereoGest} classified eight hand gestures via ANN-to-SNN conversion and demonstrated a  $3\times$ reduction in computational cost in favor of the SNN implementation.

\subsubsection{Robotic Gesture Control}
SNN-based gesture recognition has also been explored in the context of robotic control, where timely and energy-efficient decoding of human motor intent is essential. These systems typically use sEMG signals to infer gestures or motor force, translating them into actuation commands for robotic limbs or interfaces.

Several works adopt conventional approaches that convert continuous sEMG signals into spike trains via heuristic encoding schemes. For instance, Tieck et al.~\cite{tieck2020spiking} use stochastic population encoding and a layered SNN model in Nengo to decode finger-level muscle activity for controlling a five-finger robotic hand. Similarly, Scrugli et al.~\cite{scrugli2024real} implement an FPGA-based SNN pipeline that uses delta modulation to convert sEMG into spikes for both discrete gesture classification and continuous force tracking. Despite differences in architecture and hardware (e.g., CPU vs. FPGA), both systems demonstrate the feasibility of low-latency neuromorphic gesture control, with Scrugli et al. achieving latencies below 4 ms and sub-6mW average power consumption. 



In contrast, Tanzarella et al.~\cite{tanzarella2023neuromorphic} move away from artificial spike encoding by directly extracting spike trains from spinal motor neurons using high-density EMG and the Convolution Kernel Compensation (CKC) algorithm. These biologically derived spike signals are fed into a convolutional SNN trained with local learning rules (DECOLLE), yielding high classification accuracy ($\sim 95\%$
) across a range of hand gestures. This approach emphasizes biological plausibility and demonstrates the promise of preserving native spiking representations for enhanced recognition.

\subsubsection{Sign Language Gestures}

SNNs have also shown promise in recognizing structured and semantically rich hand gestures such as those used in sign language. 
These gestures often require fine-grained spatial and temporal discrimination, making them a compelling testbed for evaluating neuromorphic performance across sensing modalities and hardware platforms.

Ceolini et al.~\cite{ceolini2020hand} present a comprehensive benchmark for SNN-based sign language recognition, integrating multimodal sensing and deploying models on multiple neuromorphic hardware backends. Using both a Myo armband (for sEMG) and a DVS camera (for event-based vision), the authors classify five distinct sign language gestures using two architectures: a spiking CNN (sCNN) trained via SLAYER for deployment on Intel's Loihi, and a spiking MLP (sMLP) derived through ANN-to-SNN conversion for deployment on ODIN + MorphIC. Both setups achieve competitive accuracy -- 96\% for the sCNN and 89.4\% for the sMLP—while maintaining ultra-low energy consumption (1.1 $mJ$ and 37 $\mu J$, respectively).


\subsubsection{Navigation Control Gestures}
Beyond bio-signals and vision-based inputs, SNNs have also been integrated into non-invasive sensing modalities such as radar, expanding the applicability of neuromorphic approaches in gesture recognition.

Zheng et al.~\cite{10.1145/3625687.3625788} propose a fully neuromorphic radar sensing platform that mimics biological sensory systems to enable ultra-low power gesture recognition for IoT applications. The system employs a self-injection locking (SIL) oscillator to replace conventional RF front-ends, achieving active power consumption below 300 $\mu W$. Motion-induced perturbations are directly encoded into spike trains using an analog leaky integrate-and-fire (LIF) neuron circuit, which eliminates the need for digital preprocessing. These spike sequences are then processed by multi-layer convolutional SNNs to classify a set of 12 predefined gestures—including swipes, pushes, pulls, and static hand poses.

This end-to-end neuromorphic pipeline achieves a classification accuracy of 94.58\% and a localization error of just 0.98 m for moving targets, while cutting overall system power consumption by one to two orders of magnitude compared to traditional radar-based approaches. The work demonstrates the feasibility of combining analog spike encoding with spiking neural inference for real-time, energy-efficient gesture sensing in embedded and edge applications.

\begin{table}[t]
\centering
\resizebox{\textwidth}{!}{%
\begin{tabular}{|l|l|l|l|l|}
\hline
\rowcolor[HTML]{CBCBCB} 
\textbf{Paper} & \textbf{Spike Encoding} & \textbf{Training Method} & \textbf{Dataset} & \textbf{Accuracy} \\
\hline

\rowcolor[HTML]{F9F9F9} 
Donati et al. \cite{8747378} & Delta Modulation & Spike-based Delta Rule & EMG Myo Armband (Self-collected)& 74\%\\
\hline

\rowcolor[HTML]{FFFFFF} 
Ma et al. \cite{9073810} & Delta Modulation & STDP & EMG Myo Armband (Self-collected)   & 77\% \\
\hline 

\rowcolor[HTML]{F9F9F9} 
Vitale et al. \cite{9849452}  & Delta Modulation & SLAYER backprop & NinaPro DB5 \cite{atzori2012building} & 74\% \\
\hline

\rowcolor[HTML]{FFFFFF} 
Zhang et al. \cite{10.1016/j.ins.2021.11.065}  & Rate (poisson) & ANN‑to‑SNN & 
\begin{tabular}[c]{@{}l@{}}NinaPro DB1 \cite{atzori2012building} \\ NinaPro DB2 \cite{atzori2012building}  \end{tabular}
 & \begin{tabular}[c]{@{}l@{}} 75.5\% \\ 74.2\%  \end{tabular} \\
\hline

\rowcolor[HTML]{F9F9F9} 
Guo et al. \cite{guo2024spgesture} & ConvLIF & BPTT & sEMG gesture dataset (Self-collected)  & 89.26\% \\
\hline

\rowcolor[HTML]{FFFFFF} 
Arun et al. ~\cite{echoWrite} & Rate (poisson) & ANN-to-SNN  & Ultrasound finger gestures (Self-collected)  & 92\% \\
\hline

\rowcolor[HTML]{F9F9F9} 
Andrew et al. ~\cite{stereoGest}  & Rate (poisson) & ANN-to-SNN  & Ultrasound hand gestures (Self-collected)  & 94\% \\
\hline

\rowcolor[HTML]{FFFFFF} 
Tieck et al. \cite{tieck2020spiking}  & Rate (possion) & ANN-to-SNN Comversion & EMG Myro Armband (Self-collected) & 76.5\% \\
\hline 

\rowcolor[HTML]{F9F9F9} 
Scrugil et al. \cite{scrugli2024real}  & Delta Modulation & SLAYER Backprop &  NinaPro DB5 \cite{pizzolato2017comparison}   & 83.17\% \\
\hline 

\rowcolor[HTML]{FFFFFF} 
Tanzarella et al. \cite{tanzarella2023neuromorphic} & HD-sEMG decomposition & DECOLLE & HD-sEMG data (Self-collected)    & 95\% \\
\hline 

\rowcolor[HTML]{F9F9F9} 
Ceolini et al. \cite{ceolini2020hand}  & Delta Modulation &  \begin{tabular}[c]{@{}l@{}}  SLAYER (sCNN) \\ ANN-to-SNN (sMLP) \end{tabular}  & EMG-DVS (Self-collected) & \begin{tabular}[c]{@{}l@{}}  96\% \\ 89.4\% \end{tabular} \\
\hline

\rowcolor[HTML]{FFFFFF} 
Zheng et al. \cite{10.1145/3625687.3625788}  & Rate  & ANN-to-SNN & 12 hand gestures (Self-collected)   & 94.58\% \\
\hline 

\end{tabular}
}
\caption{Summary of SNN-based gesture recognition works.}
\label{tab:gesture_recognition}
\end{table}

\subsubsection{\textbf{Key Takeaways}} Table~\ref{tab:gesture_recognition} presents a summary of SNN-based gesture recognition works. SNNs for gesture recognition have been deployed on energy‑efficient neuromorphic chips like Loihi and low-power edge accelerators such as DYNAP‑SE, demonstrating significant reductions in latency and power consumption. However, evaluations on complex benchmarks such as NinaPro DB reveal only moderate accuracy—particularly as the gesture classes increases—highlighting the challenge of devising spike‑encoding schemes that can extract more distinctive signal features to boost multi‑class performance. Moreover, unsupervised, online learning paradigms remain largely untapped---by enabling models to adapt continuously to new users and contexts, they could markedly reduce misclassifications in real‑world scenarios. Complementing these approaches with context‑awareness mechanisms to suppress unlikely gestures would further enhance robustness. Looking forward, broadening the exploration on event‑based gesture streams with noisy electromyography (EMG) data through multimodal fusion models offers a promising direction for understanding and optimizing performance metrics across diverse and complex sensing environments. 
 

\subsection{Touch Classification}



As robots increasingly interact with unstructured environments, the ability to interpret touch becomes critical for adaptive manipulation, context-aware behavior, and fluid human-robot interaction. Real-time tactile perception, however, poses significant challenges due to high data throughput, multimodal fusion with vision, and the complex temporal dynamics of tactile signals.
While conventional deep learning methods are capable of processing such data, SNNs offer unique advantages through their event-driven, energy-efficient computation and ability to selectively encode salient features. These characteristics make SNNs particularly well-suited for tasks like reactive gripping, where fast and robust interpretation of combined visual and tactile feedback is essential.

Recent neuromorphic approaches to touch classification can be broadly categorized into five areas: (i) texture classification, (ii) touch and slip detection, (iii) multi-class touch classification, (iv) shape recognition, and (v) object and container classification. We summarize representative work in each category below.




\subsubsection{Texture Classification}

Early works in neuromorphic tactile sensing explored bio-inspired encoding schemes for texture classification but did not integrate SNNs as the primary inference mechanism \cite{rongala2015neuromorphic, rasouli2018extreme, 10.1016/j.neucom.2017.03.025}. More recent approaches have adopted SNNs explicitly, such as a fast texture classification method that utilized spike-based tactile encoding across two benchmark datasets, Biotac~\cite{gao2020supervised} and RoboSkin~\cite{taunyazov2019towards}. Using K-threshold encoding~\cite{taunyazov2020fast} and training with the SLAYER framework~\cite{shrestha2018slayer}, the SNN achieved 94.6\% and 92.2\% accuracy, outperforming traditional LSTM and SVM models. A key advantage was the system’s ability to make early decisions by emitting spikes once sufficient input information was detected, demonstrating the temporal efficiency of SNNs in dynamic tactile scenarios.


\subsubsection{Touch and Slippage Detection}

Works on touch and slip detection leverage biologically inspired SNNs and neuromorphic hardware to enable efficient, low-power tactile perception in robotic systems. One such work~\cite{follmann2024touch} proposes an SNN model simulating slow- and fast-adapting mechanoreceptors using Force Sensitive Resistors (FSRs), with input signals derived from a multi-axis JR3 force-torque sensor capturing normal and tangential forces. The network achieves high accuracy across input neuron configurations (F1~$>$0.999 for touch detection; recall between 0.994–0.998), with even small networks (e.g., 20 neurons) performing on par with larger ones. Energy profiling shows that Loihi consumes 32.66 $nJ$ per inference, while ARM and SpiNNaker2 require 0.541 $\mu J$ and 0.75 $\mu J$, respectively; in contrast, CPU execution demands 5.169 $\mu J$. The slip detection task yields similarly high F1 scores and energy profiles. Related work~\cite{Taunyazov-RSS-20} employs the NeuTouch sensor for rotational slip classification, achieving 91\% accuracy, further illustrating the suitability of SNNs for real-time tactile decision-making.

\subsubsection{Multi-class Touch Classification} 
Works on multi-class touch modality classification extend beyond basic touch detection by addressing complex tactile interactions such as press, poke, grab, squeeze, push, and wheel rolling. One work \cite{9665453} employs a piezoresistive sensor array to capture analog tactile signals, which are converted into spike trains using a Leaky Integrate-and-Fire (LIF) encoding layer. A three-layer SNN trained via a supervised STDP learning rule achieves an overall classification accuracy of 88.3\%, with most individual touch categories exceeding 90\%. In contrast, another work~\cite{sun2017categories} applies Principal Component Analysis (PCA) and a traditional k-Nearest Neighbors (k-NN) classifier to a similar sensor setup for seven touch types, reaching a lower overall accuracy of 71.4\%. These comparisons highlight the potential of SNN-based pipelines in capturing nuanced tactile dynamics for real-time classification.


\subsubsection{Shape Recognition}
Some studies have explored shape recognition using SNNs. Kim et al. \cite{9283337} collected shape data through a robotic grasping experiment using a tactile sensor array. They converted tactile signals into spikes using the Izhikevich model and processed them with an SNN for shape recognition. Their model, trained using Hebbian learning-based STDP with lateral inhibition in the second layer, achieved 100\% shape recognition accuracy. Dabbous et al. \cite{9937733} extended this study by using the same dataset but implementing a two-layer SNN trained with supervised STDP. By increasing the number of tactile sensors, they maintained 100\% accuracy while expanding classification from three to eleven objects, demonstrating the scalability of SNNs for multi-class classification with minimal computational overhead.

\subsubsection{Object Recognition and Container Classification}
Object recognition and container classification have also been widely explored using spiking graph neural networks and multimodal sensor fusion techniques. Gu et al. \cite{Gu9341421} introduced TactileSGNet, a spiking graph neural network for tactile object recognition. They constructed a tactile graph by mapping taxels to graph nodes and configured tactile geometry using manual, k-Nearest Neighbors, and Minimum Spanning Tree distance methods. The tactile graph was processed using a topology-enabled adaptive graph convolutional layer along with Leaky Integrate-and-Fire and fully connected layers. The model was evaluated on the EvTouch-Objects and EvTouch-Containers datasets, achieving 89.44\% accuracy on EvTouch-Objects and 64.4\% on EvTouch-Containers. It outperformed MLP, GCN, and grid-based CNNs, where the best-performing CNN achieved 88.40\% and 60.17\%, respectively.

Some studies have focused on optimizing SNN architectures to improve tactile object classification performance and computational efficiency. Yang et al. \cite{yang2024spiking} addressed issues of overfitting and underconvergence in SNNs by optimizing membrane potential expression and refining the loss function-parameter relationship. Their approach improved classification performance, achieving 73.33\% accuracy on EvTouch-Containers (a 4.16\% increase) and 93.75\% on EvTouch-Objects (a 2.71\% increase) compared to existing models. The model also reduced training time by 8.00\% and testing time by 8.14\%. Yang et al. \cite{10313258} further introduced AM-SGCN, an adaptive multichannel spiking graph convolutional network, to address irregular tactile sensing structures. The model constructed tactile graphs using k-Nearest Neighbors and Minimum Spanning Tree for feature aggregation and classification enhancement. AM-SGCN incorporated temporal spike learning with backpropagation to handle the non-differentiability in SNN training, achieving 91.32\% accuracy and 92.43\% precision on EvTouch-Objects.

Other research has integrated visual-tactile fusion for enhanced perception. Taunyazov et al. \cite{Taunyazov-RSS-20} developed a visual-tactile perception system for container classification and rotational slip detection using NeuTouch and a Prophesee event camera. NeuTouch, a 39-taxel graphene-based piezoresistive thin-film sensor, was designed for event-driven tactile sensing. They proposed visual-tactile SNNs trained using SLAYER backpropagation. Their setup, using a Franka Emika Panda arm with a Robotiq 2F-140 gripper, collected multimodal data from NeuTouch, Prophesee Onboard, RGB cameras, and Optitrack motion capture. Evaluated on container classification and slip detection, the model achieved 81\% and 100\% accuracy, outperforming CNN-3D (80\%) and MLP-GRU (100\%). The system was further evaluated on Intel's Loihi neuromorphic chip, demonstrating 1,900 times lower power consumption than a GPU. Wu et al. \cite{10697100} proposed VT-SGN, a visual-tactile spiking graph network for object recognition, container classification, and slip detection. The model integrated tactile and visual neuromorphic data and consisted of three components: a tactile spiking graph, a visual spiking graph, and a fusion layer for multimodal perception. The network was trained using spatio-temporal backpropagation and evaluated on object and container datasets, as well as the slip dataset, achieving 98\%, 87.2\%, and 100\% accuracy, respectively.

To improve generalization in tactile object recognition, researchers have also explored different spiking graph neural network architectures. Yu et al. \cite{10529973} and Yang et al. \cite{yang2024ggt} developed spiking graph neural networks with their models G2T-SNN and GGT-SNN, integrating Gaussian priors for tactile object recognition generalization. G2T-SNN used topological sorting and R-tree methods, while GGT-SNN incorporated M-tree and Z-tree for high-dimensional tactile data processing. GGT-SNN outperformed G2T-SNN in generalization and accuracy, achieving 75.00\% on EvTouch-Containers and 92.36\% on EvTouch-Objects, compared to 71.67\% and 91.67\% for G2T-SNN. While GGT-SNN had better spatial connectivity and generalization, G2T-SNN's topological sorting was more computationally efficient.

Kang et al. \cite{kang2023boost} have focused on the integration of spatial dependencies for enhanced tactile perception. They introduced location spiking neurons, specifically the location spike response model and the location leaky integrate-and-fire model. The location spike response model updates membrane potential based on location, while the location leaky integrate-and-fire model adapts leaky integrate-and-fire functions for spatial dependencies. Using these models, they proposed Hybrid SRM FC and Hybrid LIF GNN. The models achieved 91\% on Objects-v1, 86\% on Containers-v1, and 100\% on slip detection. Hybrid LIF GNN, representing the state-of-the-art in event-driven tactile learning, achieved 96\% on Objects-v1, 90\% on Containers-v1, and 100\% on slip detection.

\begin{table}[t]
\centering
\resizebox{\textwidth}{!}{%
\begin{tabular}{|l|l|l|l|l|}
\hline
\rowcolor[HTML]{CBCBCB} 
\textbf{Paper} & \textbf{Spike Encoding} & \textbf{Training Method} & \textbf{Dataset} & \textbf{Accuracy} \\
\hline

\rowcolor[HTML]{F9F9F9}
Taunyazov et al. \cite{taunyazov2020fast} & \begin{tabular}[c]{@{}l@{}}K-threshold Encoding \\ (Similar to LC Sampling)\end{tabular}   & SLAYER Backprop & \begin{tabular}[c]{@{}l@{}}Biotac \cite{gao2020supervised} \\ RoboSkin \cite{taunyazov2019towards} \end{tabular} & \begin{tabular}[c]{@{}l@{}}94.6\% \\ 92.2\%\end{tabular} \\
\hline

\rowcolor[HTML]{FFFFFF}
Follmann et al.  \cite{follmann2024touch} & \begin{tabular}[c]{@{}l@{}} LIF \end{tabular} & Backpropagation & FSR Dataset (Self-collected)& \begin{tabular}[c]{@{}l@{}} 99.9\% (touch) \\ 100\% (slip)\end{tabular} \\
\hline

\rowcolor[HTML]{F9F9F9} 
Dabbous et al. \cite{9665453} & LIF & Supervised-STDP & Touch Dataset (Self-collected) & 88.3\% \\
\hline

\rowcolor[HTML]{FFFFFF}
Kim et al. \cite{9283337} & Izhikevich model & WTA-STDP  & \begin{tabular}[c]{@{}l@{}} Shape Data (Self-collected) \end{tabular} & \begin{tabular}[c]{@{}l@{}} 100\% \end{tabular} \\
\hline

\rowcolor[HTML]{F9F9F9}
Dabbous et al. \cite{9937733} & LIF & Supervised-STDP & \begin{tabular}[c]{@{}l@{}}  Shape Data (Self-collected) \end{tabular} & \begin{tabular}[c]{@{}l@{}} 100\% \end{tabular} \\
\hline

\rowcolor[HTML]{FFFFFF}
Taunyazov et al. \cite{Taunyazov-RSS-20}& \begin{tabular}[c]{@{}l@{}} Tactile Electrical Pulses \\ (NeuTouch Spikes) \end{tabular}  & SLAYER Backprop & \begin{tabular}[c]{@{}l@{}}EvTouch-Objects, \\ EvTouch-Containers \cite{Taunyazov-RSS-20}\end{tabular} & \begin{tabular}[c]{@{}l@{}}81\% \\ 100\%\end{tabular} \\
\hline

\rowcolor[HTML]{F9F9F9}
Gu et al. \cite{Gu9341421} & NeuTouch Spikes & \begin{tabular}[c]{@{}l@{}}BPTT \\ (rectangular approx)\end{tabular} & \begin{tabular}[c]{@{}l@{}}EvTouch-Objects, \\ EvTouch-Containers \cite{Taunyazov-RSS-20} \end{tabular} & \begin{tabular}[c]{@{}l@{}}89.44\% \\ 64.17\%\end{tabular} \\
\hline

\rowcolor[HTML]{FFFFFF}
Yang et al. \cite{yang2024spiking} & \begin{tabular}[c]{@{}l@{}}NeuTouch Spikes\end{tabular} & \begin{tabular}[c]{@{}l@{}}BPTT (different approx) \\ with refined loss and membrane \\ potential representation\end{tabular} & \begin{tabular}[c]{@{}l@{}}EvTouch-Objects, \\ EvTouch-Containers \cite{Taunyazov-RSS-20} \end{tabular} & \begin{tabular}[c]{@{}l@{}}93.75\% \\ 73.33\%\end{tabular} \\
\hline

\rowcolor[HTML]{F9F9F9}
Kang et al. \cite{kang2023boost} &  \begin{tabular}[c]{@{}l@{}} NeuTouch Spikes \end{tabular} & \begin{tabular}[c]{@{}l@{}} SLAYER backprop (for Location SRM), \\ STBP (For Location LIF)\end{tabular}  & \begin{tabular}[c]{@{}l@{}}EvTouch-Objects, \\ EvTouch-Containers \cite{Taunyazov-RSS-20} \\ Self-Collected Slip Data\end{tabular}  & \begin{tabular}[c]{@{}l@{}}96\%  \\ 90\% \\  100\%  \end{tabular} \\
\hline

\rowcolor[HTML]{FFFFFF}
Yu et al. \cite{10529973}  & \begin{tabular}[c]{@{}l@{}} NeuTouch Spikes \end{tabular} & \begin{tabular}[c]{@{}l@{}} STBP \end{tabular} & \begin{tabular}[c]{@{}l@{}}EvTouch-Objects, \\ EvTouch-Containers \cite{Taunyazov-RSS-20}\end{tabular} &    \begin{tabular}[c]{@{}l@{}}91.67\%  \\ 71.67\%\end{tabular}  \\
\hline

\rowcolor[HTML]{F9F9F9}
Yang et al. \cite{10313258} & NeuTouch Spikes & \begin{tabular}[c]{@{}l@{}} TSSL-BP \end{tabular} & \begin{tabular}[c]{@{}l@{}} EvTouch-Objects \cite{Taunyazov-RSS-20} \end{tabular} & 91.32\%\\
\hline

\rowcolor[HTML]{FFFFFF}
Yang et al. \cite{yang2024ggt} & NeuTouch Spikes & STBP & \begin{tabular}[c]{@{}l@{}}EvTouch-Objects, \\ EvTouch-Containers \cite{Taunyazov-RSS-20} \end{tabular}  & \begin{tabular}[c]{@{}l@{}} 92\% \\ 75\%\end{tabular} \\
\hline

\rowcolor[HTML]{F9F9F9}
Wu et al. \cite{10697100} & NeuTouch Spikes & STBP & \begin{tabular}[c]{@{}l@{}} EvTouch-Objects, \\ EvTouch-Containers, \\ Slip Detection Data \cite{Taunyazov-RSS-20} \end{tabular} & \begin{tabular}[c]{@{}l@{}}  98\% \\ 87.2\% \\100\% \end{tabular} \\
\hline

\end{tabular}
}
\caption{Summary of SNN-based touch classification works. }
\label{tab:touch_classification}
\end{table}

\subsubsection{\textbf{Key Takeaways}}
Recent advances in SNNs and neuromorphic systems have shown promising results in tactile sensing tasks such as texture and touch classification, slip detection, and object recognition, as summarized in Table~\ref{tab:touch_classification}. These systems benefit from the event-driven nature of SNNs, offering both high accuracy and low power consumption, making them particularly well-suited for real-time applications in robotics and human-robot interactions. Neuromorphic platforms, such as Intel’s Loihi chip, have further demonstrated significant energy efficiency improvements over traditional computing architectures. However, several challenges remain. One major hurdle is ensuring the generalization of models across diverse environments and sensor configurations. While SNNs excel in specific tasks, they may struggle with more complex scenarios involving dynamic conditions or varying input types. Another challenge is the integration of multimodal data from different sensors, which requires sophisticated methods to ensure seamless and effective fusion. Moreover, while these systems have shown promising results in controlled experiments, real-time processing and scalability remain concerns for deployment in more complex or large-scale systems. These challenges present opportunities for further research into more robust SNN architectures, improved multimodal fusion techniques, and approaches that can handle a wider range of real-world conditions.

%% file: sections/05_hardware_software.tex
\section{Software and Hardware Platforms}

ANNs have extensive software support. Tools like PyTorch and Tensorflow offer comprehensive, well-maintained libraries that are regularly updated with state-of-the-art models and come with good documentation. The same is true for hardware support, as most hardware platforms are standardized with well-described system architectures and clear specifications, whether they are GPU/CPU-based or edge-based microcontrollers. In contrast, support for SNNs is still emerging. Various functionalities are distributed across different software and hardware tools: some software is designed solely for simulation, while others serve as broader development frameworks. Moreover, the training support provided by SNN software varies significantly along with the neuron models they support, and only a few platforms offer direct integration with neuromorphic hardware. Furthermore, several neuromorphic hardware platforms have recently emerged in the realm of low-power neuromorphic processing; these are particularly useful for deploying resource-constrained models for ubiquitous applications. 

Motivated by these gaps, this section presents a curated overview of well-maintained SNN software and hardware, highlighting their key features. \textit{This overview aims to support researchers interested in pursuing SNN research by helping them select appropriate tools for developing and deploying models based on their specific needs.}

\begin{table}[h!]
\centering
\resizebox{\textwidth}{!}{%
\begin{tabular}{|l|p{3cm}|p{3cm}|p{3cm}|p{3cm}|}
\hline
\textbf{Software} & \textbf{Type} & \textbf{Neuron Models} & \textbf{Training Support} & \textbf{Neuromorphic Hardware Support} \\ \hline

BindsNET \cite{hazan2018bindsnet} & Simulation Software & LIF, IF, McCulloch Pitts, Izhikevich & STDP/RSTDP & No support \\ \hline

Brian2 \cite{stimberg2019brian} & Simulation Software & Hodgin-Huxley , LIF & STDP & No support \\ \hline

GeNN \cite{yavuz2016genn} & Simulation Software & LIF, Izhikevich, Possion & STDP & No support for neuromorphic, but NVIDIA GPUs.  
\\ \hline

NEST \cite{Gewaltig:NEST} & Simulation Software & LIF, IF, Hodgkin-Huxley and a wide range of models & STDP, custom training rules & No neuromorphic support but NVIDIA GPUs \\ \hline

snnTorch \cite{eshraghian2023training} & Deep Learning Framework &  LIF, rLIF, Lapicque, Alpha, Spiking LSTM, Spiking 2d-conv LSTM  & Supervised (Gradient based) - BPTT & No support \\ \hline

SpikingJelly \cite{fang2023spikingjelly} & Deep Learning Framework & IF, LIF, Izhikevich  & STDP, ANN-to-SNN & Supports conversion to Lava (Loihi 2) \\ \hline

Nengo \cite{bekolay2014nengo} & Deep Learning Framework & LIF, Izhikevich& PES, RLS, BCM & Intel Loihi, SpiNNaker \\ \hline

Norse \cite{norse2021} & Deep Learning Framework & LIF, IF, Izhikevich  & STDP, SLAYER, Surrogate-Gradients, Tsodyks-Markram learning & No support \\ \hline

Lava/Lava-DL \cite{lava2024} & Deep Learning Framework & LIF, R\&F Izhikevich, Adaptive LIF, Sigma Delta & STDP, SLAYER, ANN-to-SNN & Intel Loihi 2 and NVIDIA GPUs \\ \hline

Sinabs \cite{sinabs} & Deep Learning Framework & LIF, IF, Adaptive LIF, Exp LIF & BPTT, ANN-to-SNN & Speck, DYNAP-CNN \\ \hline

Rockpool \cite{muir_dylan_2019_4639684} &  Deep Learning Framework & LIF & Gradient-decent (BPTT) & Xylo, DYNAP-SE2 \\ \hline
\end{tabular}%
}
\caption{Summary of neuromorphic software tools.}
\label{fig:software_table}
\end{table}

\subsection{Software Platforms}

Neuromorphic software platforms remain under-described and insufficiently categorized in the existing literature. In this subsection, we outline key tools and their features—focusing on software types (simulation vs. development libraries), support for neuron models, training strategies, and compatible hardware. Table \ref{fig:software_table} provides a comparative summary of both established and emerging platforms. We then present a detailed overview of each.
\begin{enumerate}
    \item \textbf{BindsNET} \cite{hazan2018bindsnet} is a Python-based simulation software (released in 2018) for spiking neural networks that leverages PyTorch tensors for computing on CPUs or GPUs. While it does not offer neuromorphic hardware support, the simulation library provides a variety of functionalities for machine learning and reinforcement learning applications. It includes collections of neuron models—such as LIF, IF, McCulloch-Pitts, and Izhikevich—with training methodologies supported by STDP and reward-modulated STDP. Moreover, it is equipped with datasets like MNIST, CIFAR-10, CIFAR-100, as well as RL-based environments such as GymEnvironment. 

    \textit{Comments on support:} The software is readily usable and can be installed using pip. Its documentation is minimal, just two pages covering how to add network components and learning rules, but it is well-written and the API is clearly structured. While sufficient for simulating and monitoring the state variables of SNNs, its scope is primarily toward computer vision tasks. 

    \textit{Link for documentation and further reading:} \url{https://bindsnet-docs.readthedocs.io/}

    \item \textbf{Brain2} \cite{stimberg2019brian} is an open source simulation software (released in 2013) for spiking neural networks. It supports Hodgkin-Huxley and LIF neuron models and allows for the definition of synapses, which are complemented with STDP training. Its focus is on simple implementations using mathematical modeling, making it a beginner's choice for trying neuromorphic simulations. Brain doesn't have any hardware support or built-in datasets. Link for documentation and further reading: \url{https://briansimulator.org/}

     \item \textbf{GeNN} \cite{yavuz2016genn} is a simulation software (released in 2014) that accelerates implementations on hardware, particularly for NVIDIA GPUs and CPUs. It uses a C/C++ backend to port models to NVIDIA's CUDA. The SNN models are built using the Python API and support typical neuron models such as LIF, Izhikevich, and the Poisson model with synaptic connections. These models can be trained using STDP learning rules. Link for the documentation and further reading: \url{https://genn-team.github.io/}

     \item \textbf{NEST} \cite{Gewaltig:NEST} is a simulation software (released in 2007) for spiking neural networks that supports a wide range of neuron models, including LIF, IF, adaptive LIF, the MAT2 model, and Hodgkin-Huxley. It supports synaptic plasticity mechanisms such as short-term plasticity, STDP, and static synaptics, along with customizable connection rules. It is specifically designed to handle large networks with millions of synapses and neurons. The interpreter uses multi-threading for efficient processing and handling of larger models. Furthermore, NESTGPU enables these large-scale models to run on NVIDIA GPUs, which can be used with Python/C/C++. Link for the documentation and further reading: \url{https://www.nest-simulator.org/}
    

    
    \item \textbf{snnTorch} \cite{eshraghian2023training} is a deep learning framework (released in 2021) that adapts SNN models and can be used in conjunction with PyTorch. It supports various LIF-based neuron models, including recurrent LIF, leaky integrate-and-fire models, and second-order integrate-and-fire neurons. Further, it's spikegen module has rate, latency and delta encoding schemes to convert real valued data into spike trains. Models constructed using snnTorch are trained using BPTT techniques and currently support only supervised learning. Additionally, it includes neuromorphic datasets such as DVS-Gestures, NMNIST, and the SHD dataset, which can be used alongside the data processing toolkit Tonic \cite{lenz_gregor_2021_5079802}. This helps to load and manipulate spike-based data. 

    \textit{Comments on support:} This framework is quite user-friendly, with pip installation and tutorials featuring sample code chunks that help beginners get started quickly. Its documentation is comprehensive, enhanced with clear illustrations, plots, and a well-defined API structure. These features make it sufficient for developing, visualizing, and evaluating SNN models. One can develop both vision-based and time-series based application in snnTorch.

    \textit{Link for documentation and further reading:} \url{https://snntorch.readthedocs.io/}

    \item \textbf{SpikingJelly} \cite{fang2023spikingjelly} is an open-source deep learning framework (released in 2019) with extensive support for SNN pipelines. It offers various neuron models, including IF, LIF, and their variants such as Quadratic LIF, Exponential LIF, Gated LIF, and the Izhikevich model. It supports training using STDP and provides conversion capabilities from ANN to SNN through PyTorch. Furthermore, it extends its support by allowing SNN models to be converted to the Lava Framework \cite{lava2024}, enabling deployment on the Intel Loihi 2 chipset. It is equipped with a variety of image neuromorphic datasets, such as N-MNIST, CIFAR10-DVS, DVS128 Gesture, N-Caltech101, and ASLDVS. It also supports large, standard torch-based models in the spiking domain, such as Spiking ResNet. 

    \textit{Comments on support:} This framework is highly accessible, with pip installation. While its tutorials aren’t as comprehensive as those offered by snntorch, the documentation is clear and robust. It supports a wider variety of neurons and offers features like unsupervised learning and ANN-to-SNN conversions, contributing to efficient training processes. Its has modules supporting  applications in speech, vision, gestures, and broader image/time-series analysis.

    \textit{Link for documentation and further reading:} \url{https://spikingjelly.readthedocs.io/}
    

    \item \textbf{Nengo} \cite{bekolay2014nengo} is a Python-based, open-source deep learning framework (released in 2015) with various deployment and simulation libraries. It supports basic neuron models, such as the LIF and Izhikevich models, and extends them with variants like the Adaptive LIF. It also supports synapse models, including the linear filter synapse and the alpha filter synapse. Additionally, it enables running Nengo models on Intel's Loihi architecture as well as on SpiNNaker, using dedicated backends—NengoLoihi and NengoSpiNNaker. It uses learning rules like Prescribed Error Sensitivity (PES), Recursive least-squares rule (RLS) and Bienenstock-Cooper-Munroe (BCM). 
    Link for documentation and further reading: \url{https://www.nengo.ai/nengo/}

    \item \textbf{Norse} \cite{norse2021}  is another deep learning framework (released in 2020) which extends PyTorch’s functionalities to support and develop spiking neural networks (SNNs). It employs the Leaky Integrate-and-Fire (LIF) model and its variants as well as the Izhikevich model, with support for plasticity training via STDP and Tsodyks–Markram synaptic dynamics. Additional training methods, such as SuperSpike and surrogate gradient learning, are also provided. Although it does not offer neuromorphic hardware support, it includes datasets like MNIST, CIFAR, and cartpole balancing (using policy gradients) for offline development and testing. Link for the documentation and further reading: \url{https://norse.github.io/norse/}
    
    \item \textbf{Lava/Lava-DL} \cite{lava2024} is an open-source deep learning framework (released in 2021) for the neuromorphic community that provides various features and capabilities for developing SNN models and porting them to Intel's Loihi. It operates via a Python interface while also supporting low-level C/C++/CUDA blending. This enables support not only for the Loihi 2 chipset but also for other CPU and GPU-based hardware. Lava includes a core neuron model (LIF) and it's variants like adaptive LIF, Sigma delta and Resonate and Fire (RF) Izhikevich models. These can be trained using a variety of techniques---STDP, SLAYER and ANN-to-SNN conversion. Lava framework is more suitable for image processing tasks, with built-in datasets like MNIST. 
    
    
    Link for the documentation and further reading: \url{https://lava-nc.org/}
   
    \item \textbf{Sinabs} \cite{sinabs} is a PyTorch-based deep learning framework (released in 2019) for developing spiking neural networks. It is specifically designed for vision-based models. It supports LIF and its variants, such as Adaptive LIF, IF, and Exponential LIF. It uses BPTT and ANN-to-SNN conversion for training, and it is well documented with tutorials included. It comes with the MNIST dataset built in. This library is compatible with Synsense chipsets—Speck and DYNAP-CNN (also focused on vision)—and provides a detailed description of how to deploy the model on these chipsets, measuring power consumption and spike count visualization. Link for the documentation and further reading: \url{https://sinabs.readthedocs.io/}
    
    \item \textbf{Rockpool} \cite{muir_dylan_2019_4639684} is another deep learning framework (released in 2019) that caters to lower-dimensional signals with event driven dynamics, thereby enabling hardware support for Xylo and DYNAP-SE2. It also provides features such as measuring the dynamic and idle power consumption of the chipsets. Since Rockpool is designed for ultra-low-power hardware, it exclusively uses the LIF model in its framework, with gradient descent-based training methods supported by Jax and PyTorch.

      \textit{Comments on support:} This framework is easy to install using pip or conda. Its documentation includes tutorials for training models with both PyTorch and JAX backends and provides instructions for deploying models to Xylo hardware. It offers clear high-level and low-level API structures. The framework supports time-series data and ultra-low power applications, while also allowing for the training of relatively larger models with Rockpool.

    \textit{Link for documentation and further reading:} \url{https://rockpool.ai/} 
    
\end{enumerate}

\textbf{\textit{Key takeaways:}} We make the following recommendations for readers with different needs considering ubiquitous computing paradigm. 

\begin{itemize}
    \item If the reader wants to replicate the dynamics and temporal properties of neurological systems using spiking neurons and their interactions within networks, he/she should select simulation tools. Among these, \textbf{BindsNET} stands out as a top choice due to its native integration with PyTorch, support for rapid prototyping, and GPU-accelerated training of SNNs. 
    \item Readers interested in developing and modifying conventional ANN–based approaches for building SNN models and training them efficiently using techniques such as BPTT and STDP should opt for general-purpose deep learning frameworks. Among the available options, \textbf{SpikingJelly} is highly recommended due to its extensive support for various neuron models, facilitation of ANN-to-SNN conversion and STDP-based training, and compatibility with the Lava Framework for deployment on neuromorphic hardware like Intel’s Loihi2 chipset. Additionally, snnTorch is a strong alternative, especially when the target application relies on gradient-based supervised learning and requires flexible spike encoding schemes to convert digital data into time-varying spike trains.
    
    \item If the reader aims to build systems that process time-series sensor data (such as audio or IMU signals) and deploy them on low-power edge hardware should consider application-specific deep learning frameworks. Rockpool is particularly well-suited for such edge-focused deployments, offering integration with PyTorch and JAX backends, along with built-in tools for converting models into hardware-specific computational graphs compatible with platforms like Xylo and DYNAP-SE2.


    \item{Finally, while software simulators such as snnTorch and BindsNET offer powerful platforms for designing, tuning, and training SNN models for optimum performance, they do not ensure that the same SNN can be ported to any neuromorphic hardware without compromising the performance. This is largely due to the fact that neuromorphic devices often incorporate custom neuron models, connection parameters, and other architectural constraints that are specifically optimized for their hardware environment. Readers who wants to implement their models on any hardware should consider designing and developing the SNNs in software stacks that are coupled with a development board. For instance, one can use Lava for Loihi or Rockpool for Xylo. Nonetheless, seamless portability of SNNs designed on general-purpose software platform to neuromorphic hardware remains an open area of research and development.}
    
\end{itemize}








\subsection{Neuromorphic Hardware}
\label{sec:hardware}
\begin{figure}[ht]
    \centering
    \includegraphics[width=0.8\linewidth]{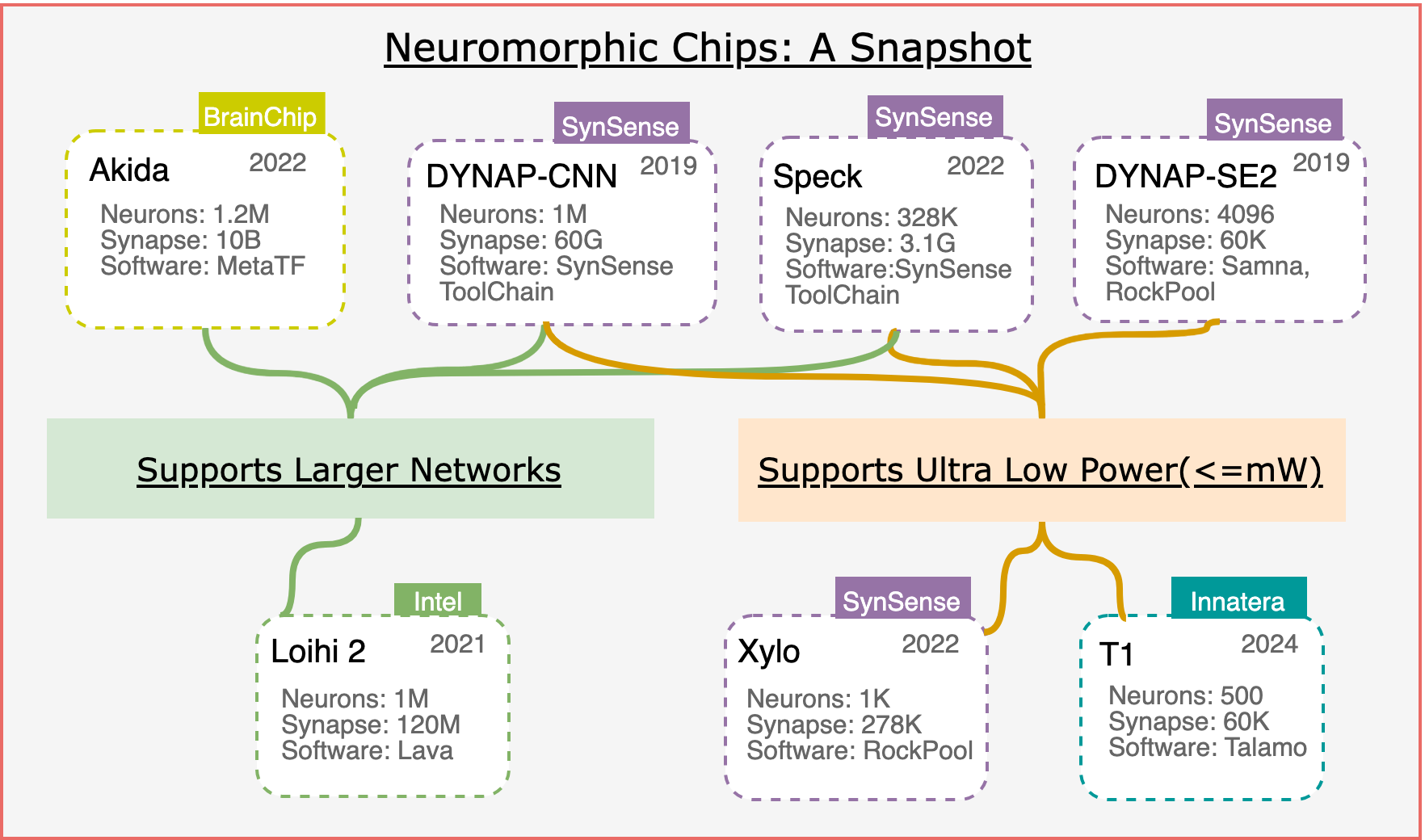} 
    \caption{An illustration of neuromorphic hardware chipsets.}
    \label{fig: Hardware}
\end{figure}


In this subsection, we summarize the latest commercially available neuromorphic hardware chipsets, highlighting their capabilities and development support for building and deploying spiking neural networks. We focus on widely used chipsets across two categories: general-purpose and application-specific (ultra-low-power) devices. Particular emphasis is placed on ultra-low-power processors, which are well-suited for time-series data that require continuous monitoring, constrained compute, and energy-efficient operation in ubiquitous applications. Figure \ref{fig: Hardware} provides an overview contrasting chipsets designed for large-scale model deployment with those optimized for ultra-low-power consumption (i.e., $\leq 1,\mathrm{mW}$). The following list presents chipsets in detail in descending order of network capacity, ending with those best suited for energy-constrained, ubiquitous applications. 



\begin{enumerate}

    \item \textbf{Akida} \cite{brainchip_akida} by BrainChip Inc. is a neuromorphic processor that can support advanced CNNs, Vision Transformers and Temporal Event-based Neural Networks (TENNs). Available chipsets include AKD1000, AKD1500, and AKD2000, although only the support and maintenance for AKD1000 is considered reliable. The first-generation AKD1000 was designed only for CNNs and RNNs, while the second-generation AKD1500 and AKD2000 support TENNs and Vision Transformers. Applications include time-series forecasting, vital sign prediction, video object recognition, LiDAR analysis, and speech processing. The chipset uses the MetaFT framework, with Python as the backbone for hardware processing and deployment. It offers a rich model zoo spanning a variety of networks and employs wrappers to covert quantized ANNs into SNNs during training.  It also supports on-chip learning---allowing the network to adapt to new data after deployment.
    

    \item \textbf{Loihi 2} \cite{davies2018loihi} by Intel is the most widely used chip for implementing SNN-based architectures on-chip. Featuring 1 million neurons and 120 million synapses, this neuromorphic research chipset is available only through the Intel Neuromorphic Research Community (INRC). Although it is not specifically optimized for low-power settings, it serves as a general-purpose platform for benchmarking model performance. Lava has been used as the software framework to build models with its Python API and deploy them to Loihi 2. Loihi chipsets have catered to a diverse range of applications—from efficient video processing to simpler human activity recognition (HAR) tasks. 
    
    
    
    \item \textbf{DYNAP-CNN} \cite{dynapcnn} by SynSense is a convolutional spiking neural network architecture–based chipset with 1 million neurons on board and over 60G synapses. It is specifically designed for capturing spatial features in vision-based tasks and can be used for ultra-low-power applications consuming less than one milliwatt of power \cite{liu2019live}. It utilizes the Synsense neuromorphic toolchain—including Sinabs and Samna—for deploying models onto the hardware. 

    \item \textbf{Speck} \cite{richter2023speck} by SynSense is a low-power neuromorphic SoC specifically designed for vision-based tasks. It supports spiking CNNs with a Dynamic Vision Sensor (DVS), featuring 328K neurons, over 6.1G synapses, and operating at approximately milliwatt power levels. It utilizes LIF neurons. Its applications include obstacle tracking and detection, driver attention tracking, sign recognition, fall detection, and gesture control. It employs the SynSense neuromorphic toolchain for training and deploying convolutional neural networks on the board. 

    \item \textbf{DYNAP-SE2} \cite{richter2024dynap} from SynSense is an ultra-low-power neuromorphic chipset with only 4,096 neurons and 65K synapses that supports recurrent networks. It uses Samna for streamlining tasks and Rockpool for model development. Its on-chip interfaces directly connect with DVS and other event-based sensors for time-series data such as audio and ECG.

    \item \textbf{Xylo} \cite{synsense_xylo} from SynSense is an ultra-low-power SNN inference chipset with only 1K LIF neurons and 278K synapses. It supports typical MLP-based SNN architectures with sensing capabilities provided through both Xylo Audio ($\sim550$ µW) and IMU-based (<300 µW) development boards. It uses Rockpool as its development and deployment platform, which is well documented and maintained. These boards are intended for very low-power or battery-powered applications, such as fall detection, human behavior detection, ambient noise classification, and keyword spotting.

    \item \textbf{T1} \cite{innatera_t1} from Innatera Nanosystems B.V. is an ultra-low-power (<1 mW) SNN accelerator with 500 neurons and 60K synapses. It is ideal for hearables, wearables, or any battery-powered, always-on sensing applications. Talamo is the software used with T1 to run SNN models and to map traditional PyTorch-based models onto the spiking neural processor chipset. It follows a PyTorch-based development flow. Typical applications demonstrated or projected include audio scene classification, gesture recognition, human presence detection, and audio sound classification.


\end{enumerate}



\textbf{\textit{Key takeaways}}: We make the following recommendations for readers with different needs considering neuromorphic hardware chipsets.
\begin{itemize}
    \item Readers seeking a research-oriented, general-purpose platform for developing and experimenting with SNN architectures should consider \textbf{Loihi 2}. This neuromorphic chipset is specifically designed for flexible and exploratory research, supporting efficient prototyping and evaluation of novel SNN models. 

    \item For application-specific deployments—especially those involving low-power, time-series processing on edge devices—specialized hardware such as \textbf{DYNAP-SE2} and \textbf{Xylo} is recommended. These platforms are optimized for resource-constrained computational models, offering reliable performance for continuous monitoring and efficient sensor data processing. Additionally, for low-power, vision-based applications, DYNAP-CNN and Speck are also strong candidates.


\end{itemize}


%% file: sections/06_future_directions.tex
\section{Future Research Directions}
While sufficient work has demonstrated the compelling advantages of SNNs, such as energy-efficient, event-driven computation, their integration into ubiquitous computing systems remains underexplored. Prior studies have demonstrated the potential of SNNs for time series tasks, ranging from gesture recognition to audio processing, but these efforts have largely focused on algorithmic feasibility in controlled settings, rather than on end-to-end systems that incorporate real-time sensing, on-device inference, and integration with heterogeneous hardware. Practical factors central to ubicomp research, such as motion artifacts, distribution shifts, and user/environmental variability, are often overlooked, limiting confidence in real-world deployment. Furthermore, while SNNs often assume access to neuromorphic inputs or employ simplified spike encoding methods, there is currently no standardized and scalable approach for converting analog sensor data into spike trains, unlike the well-established analog-to-digital conversion in conventional pipelines. This gap poses practical challenges when integrating SNNs with commonly used sensors in ubiquitous systems. However, recent advances in neuromorphic hardware and supporting software frameworks are rapidly lowering these barriers. Tools for SNN training, deployment, and debugging are becoming more accessible, and emerging platforms now support real-time, energy-efficient processing suitable for edge environments. These developments create a timely opportunity for the ubicomp community to revisit SNNs through the lens of systems thinking and explore promising directions, which we outline in the following paragraphs.


\begin{itemize}

    \item \textbf{Streaming-based Processing with SNN:} Compared to ANNs, SNNs are uniquely suited to capturing the temporal dynamics of input signals. Specifically, each SNN neuron maintains a membrane potential that accumulates over multiple sequential inputs and fires a spike once a threshold is reached. This accumulation mechanism allows SNNs to retain past states—effectively giving them a form of memory—which makes them inherently well-suited for analyzing temporal data. However, existing SNN-based approaches to time series analysis often follow the conventional ANN paradigm, where the input stream is segmented into overlapping windows (i.e., the sliding window technique) to avoid missing important events. While effective, this approach leads to redundant processing of overlapping data segments, resulting in increased computation and latency. More importantly, it undermines the temporal processing advantage of SNNs by imposing a fixed windowing structure. Therefore, a promising direction is to explore streaming-based processing pipelines, \textit{where each sensor data point is fed into the SNN individually and continuously, without windowing}. Enabling this streaming paradigm presents two key challenges: (1) Training with individual data points – since each sample represents a single point in time, it is difficult to assign meaningful labels for supervised learning without finer-gained differentiation (e.g., does a particular accelerometer reading sample at the start of a `walk' activity correspond to a walk, stand or some other intermediate activity label?)---this can complicate the training process; (2) Adjusting the memory size – different sensing tasks may require different temporal dependencies, so the network must adapt its memory to the task at hand.
    
    \item \textbf{Multi-Modal SNNs:} While the current survey has focused on single-modality SNNs, processing isolated streams such as inertial, auditory, or physiological data, there is growing interest in multi-modal SNNs, where heterogeneous sensor modalities are fused to enable richer, context-aware perception. For example, prior works~\cite{Taunyazov-RSS-20} demonstrated the fusion of event-based vision and neuromorphic touch in robotics to enable grasp stability and edge-aware object manipulation. This combination leverages the complementary timing and sparsity of visual and tactile events to improve fine motor control. Beyond vision-touch, future work could explore SNNs that integrate auditory and visual spiking inputs for applications like low-latency speechreading or audiovisual perception in noisy environments. Another promising direction is the fusion of inertial sensing with bio-signals such as EMG or ECG, allowing embodied agents to infer intent or affective state alongside motion, which is particularly relevant in assistive robotics or implicit cue-based human-robot interaction. These multi-modal SNNs open new challenges in temporal alignment, encoding schemes, and biologically plausible fusion mechanisms, but also offer the opportunity to emulate the integrative efficiency of natural sensorimotor systems.

    \item \textbf{Responsiveness of the SNNs:} Another important future direction involves improving the responsiveness of SNNs, specifically, enhancing their ability to produce accurate outputs within the first few time steps after stimulus onset. This is critical in applications such as early anomaly detection in physiological signals or low-latency audio classification. While neuron models such as  the Adaptive Leaky Integrate-and-Fire (ALIF)~\cite{bellec2018long} have introduced dynamic thresholding based on recent spiking history to enhance temporal sensitivity, these mechanisms are not yet optimized for fast, early inference when the input stream is noisy or sparse. Future work could explore training strategies that explicitly reward early correct spikes, such as time-weighted loss functions~\cite{nar2020learning} that bias learning toward rapid recognition. Additionally, input encoding methods that emphasize high-information transients can help SNNs generate meaningful responses earlier in the sequence. Together, these approaches could significantly improve the latency-performance trade-off in SNNs handling continuous sensor streams.


    \item \textbf{Novel Encoding Schemes and Hardware Implementations:} Another promising direction for future research lies in the development of novel spike-encoding schemes tailored to the rich diversity of ubiquitous sensing signals. Hybrid encoding methods—such as combining delta and rate-based schemes—could dynamically adjust intra-sample spike generation, employing higher firing rates during rapid signal fluctuations to enhance temporal resolution, and lower rates during steady-state periods to conserve energy. Entropy‑based encodings might offer another direction by generating spikes in response to shifts in signal complexity, which is especially relevant for applications like human activity recognition or early seizure detection via EEG entropy gradients (which remains underexplored). To fully exploit these advanced encoding strategies, neuromorphic front-end hardware should also evolve beyond traditional analog-to-digital converters commonly used in sensing pipelines and instead use programmable analog components (e.g.,~\cite{chen2022neuromorphichardware, zanoli2023hardware}) to generate spike trains. Event‑driven sensing circuits capable of directly extracting novel features such as phase information, correlation, and entropy would enable more expressive and multi‑channel spike representations. Integrating such circuits with programmable encoding pipelines on low-power neuromorphic chips could bridge the gap between sensor front ends and spiking processors, enabling more efficient on-device processing and substantially reducing data bandwidth requirements.

    \item \textbf{Model Adaptation in SNN:} A major research focus in the ANN domain in recent years has been adapting trained models to dynamically changing environments after deployment—an area encompassing subfields such as domain adaptation~\cite{farahani2021brief}, test-time adaptation~\cite{liang2025comprehensive}, and continual learning~\cite{de2021continual}. In these scenarios, model weights are typically fine-tuned to accommodate shifts in data distribution caused by real-world condition changes, often in an unsupervised manner due to the lack of labeled data. Various techniques have been developed, such as entropy minimization, pseudo-label generation, class-incremental learning, etc. 
    However, due to the inherent structural differences between SNNs and ANNs, existing adaptation methods cannot be directly applied to the SNN domain. A key challenge stems from the memory mechanism in SNNs, which retains temporal information across spikes during inference. This makes it particularly difficult to model how distribution shifts manifest over time. Consequently, there is a pressing need for novel approaches that enable effective and efficient adaptation in SNNs. Furthermore, since SNNs are typically deployed on dedicated neuromorphic hardware, any adaptation technique must also account for hardware constraints and execution characteristics to ensure practical applicability.


    \item \textbf{SNN Compression for Edge/End Devices:} Although most directly trained SNN models are currently small in scale, the growing trend toward larger models (e.g., Spikformer~\cite{zhou2023spikformer}) - along with the increasing feasibility of converting large ANNs to SNNs—necessitates the development of compression techniques tailored for SNN deployment on edge or resource-constrained end devices. The applicability of conventional model compression methods from the ANN domain, such as pruning, quantization, dropout, and knowledge distillation, remains largely unexplored in the context of SNNs due to the fundamental differences in their computational paradigms.  These differences call for the design of novel, SNN-specific compression strategies that account for the event-driven and temporal nature of spiking networks. As a concrete example, model compression strategies for vision ANNs have exploited the predicted \emph{spatial} saliency (e.g.,~\cite{hou2022compression}) of different channels/features; in contrast, SNN compression techniques should exploit both spatial \emph{and temporal} saliency (e.g., the timing of neural spikes).  By framing SNN compression as Neural Architecture Search (NAS) problem, we can discover architectures that minimize  parameter count and adjusts the spike activity without manual pruning. Moreover, because SNN execution is often closely tied to the characteristics of the target hardware, compression techniques should be co-designed with hardware considerations in mind, integrating hardware-in-loop cost models in the NAS ensures to maintain the on-chip energy efficiency and latency targets.  

    \item \textbf{Middleware for SNN:} Currently, neuromorphic application developers face several challenges. First, the initial learning curve is steep, requiring an understanding of neuron and synapse functionalities, learning algorithms, and computational neuroscience concepts. Second, there is significant heterogeneity across neuromorphic hardware platforms—differing in capabilities, instruction sets, neuron models, synaptic architectures, and learning paradigms—all with their own design rationales. Developers must navigate these intricacies to successfully build applications. Third, applications often require complete re-engineering to port from one hardware platform to another. These issues have been repeatedly highlighted at major academic events and forums. To address this, a standard neuromorphic middleware or software stack is needed—one that abstracts hardware complexities, similar to how the CUDA ecosystem or Intel’s OneAPI supports their respective domains. Such a middleware would allow data scientists, engineers, and developers to adopt evolving hardware platforms more easily, speeding up development, reducing reliance on niche expertise, and enhancing scalability.
    

    
    \item \textbf{Novel SNN Applications:}
    A major research thrust in the ubiquitous computing community is the exploration of novel applications enabled by innovative sensing modalities or principles. As reviewed in Section~\ref{sec:snn_applications}, current research on applying SNNs to time series data has largely focused on well-established tasks—such as human activity recognition, audio classification, and emotion recognition—primarily aiming to replace traditional ANNs with SNNs for gains in energy efficiency and latency. However, a compelling and underexplored direction is to ask whether entirely new applications could be enabled by the unique characteristics and capabilities of SNNs. One intuitive way to approach this question is by examining areas where the human brain significantly outperforms current ANNs—given that SNNs more closely mimic biological neural processes. For instance, humans excel at learning how to learn (meta-learning), generalizing from few examples, and adapting quickly to changing environments. These capabilities suggest exciting potential for SNNs in reinforcement learning tasks, particularly those involving robotic control, adaptive behavior, or autonomous systems. Exploring how SNNs can be leveraged or combined with existing reinforcement learning frameworks could open the door to more efficient, adaptive, and brain-like learning in real-world ubiquitous computing scenarios.

\end{itemize}

%% file: sections/07_conclusion.tex
\section{Conclusion}
In this survey, we have highlighted the advances, promises and challenges of SNNs to the ubiquitous systems community by analyzing 76 peer-reviewed papers across six key application domains. We also reviewed major software platforms in terms of usability, features, and functionality, and summarized neuromorphic chipsets based on application scope, power requirements, and support—offering practical guidance for SNN development. Finally, we outlined several promising research directions that we believe are relevant to the interests and needs of the ubiquitous computing community. We believe this survey can serve as a catalyst for a new wave of research, inspiring innovation and facilitating the practical deployment of SNN-based solutions in real-world ubiquitous systems.

